\PassOptionsToPackage{table}{xcolor}
\documentclass{article}
\usepackage{amssymb}

\usepackage[preprint]{corl_2025} 

\usepackage{graphicx}
\usepackage{subcaption}
\usepackage{adjustbox}
\usepackage{enumitem}
\usepackage{amsmath}
\usepackage{cleveref}
\usepackage[most]{tcolorbox}
\newtcolorbox[auto counter, number within=section]{promptbox}[2][]{colback=gray!10, colframe=black, fonttitle=\bfseries, sharp corners,title=Box~\thetcbcounter: #2, #1, label=box:\thetcbcounter}
\crefname{promptbox}{Box}{Boxes}

\usepackage{wrapfig}

\usepackage{booktabs}
\usepackage{caption}

\title{Scan, Materialize, Simulate: A Generalizable Framework for Physically Grounded Robot Planning}

\author{
  Amine Elhafsi$^1$ \hspace{0.3cm} Daniel Morton$^1$ \hspace{0.3cm} Marco Pavone$^{1,2}$ \\
  $^1$Stanford University \hspace{1cm} $^2$NVIDIA Research\\
  \texttt{\{amine, dmorton, pavone\}@stanford.edu}
}

\begin{document}
\maketitle


\begin{abstract}
    Autonomous robots must reason about the physical consequences of their actions to operate effectively in unstructured, real-world environments. We present \textbf{Scan, Materialize, Simulate} (SMS), a unified framework that combines 3D Gaussian Splatting for accurate scene reconstruction, visual foundation models for semantic segmentation, vision-language models for material property inference, and physics simulation for reliable prediction of action outcomes. By integrating these components, SMS enables generalizable physical reasoning and object-centric planning without the need to re-learn foundational physical dynamics. We empirically validate SMS in a billiards-inspired manipulation task and a challenging quadrotor landing scenario, demonstrating robust performance on both simulated domain transfer and real-world experiments. Our results highlight the potential of bridging differentiable rendering for scene reconstruction, foundation models for semantic understanding, and physics-based simulation to achieve physically grounded robot planning across diverse settings.
\end{abstract}



    \begin{figure}[h]
        \centering
        \includegraphics[width=0.825\linewidth]{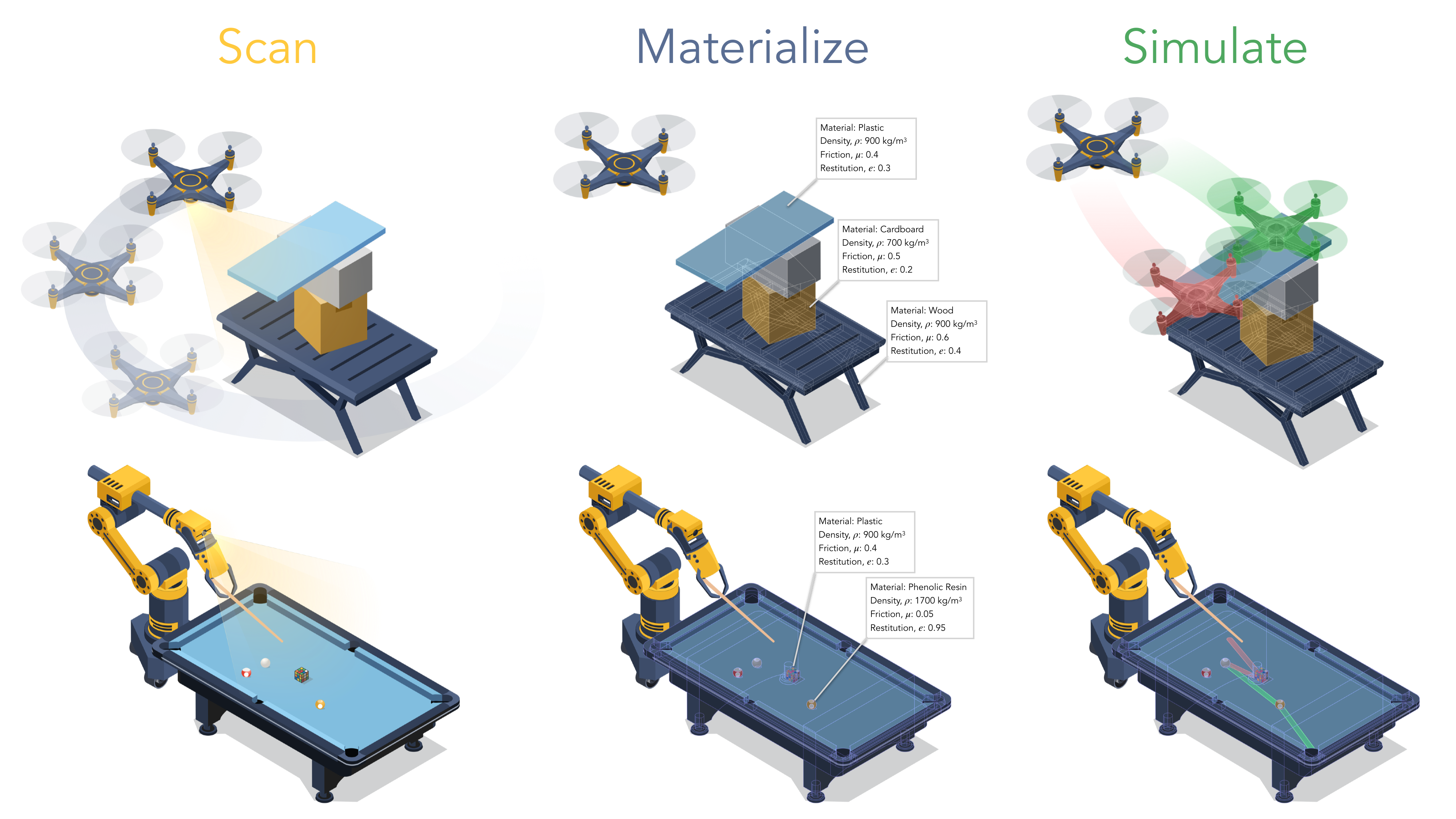}
        \caption{Overview of the \textbf{Scan, Materialize, Simulate} (SMS) framework for physics-informed robot action planning. SMS consists of three steps: scanning to build a geometric environment model, materializing to convert this reconstruction into a simulation-ready representation, and simulating to optimize actions in a virtual environment prior to target-environment execution.}
        \label{fig:hero}
    \end{figure}

\vspace{-0.1in}
\section{Introduction}
\vspace{-0.1in}
    In recent years, autonomous robotic systems have demonstrated remarkable capabilities, paving the way for deployment across a wide range of real-world applications. However, tasks that require a robot to reason about the physical consequences of its actions remain particularly challenging. Many existing systems tackle these challenges only in highly structured environments, under rigid assumptions, or with highly specialized algorithms; for example, custom solutions have been developed to enable robots to play physically challenging games such as Jenga~\citep{Fazeli2019}, billiards~\citep{greenspan2008toward, elmoutaouaffiq2025} and table tennis~\citep{d2023robotic}. Beyond such recreational settings, diverse domains motivate the need for robust reasoning about physical interactions in unstructured environments, whether a robot is negotiating unstable rubble in disaster response, stacking irregular loads in warehouses, or handling equipment in microgravity space environments. In each of these scenarios, the robot's success hinges not just on accurate control, but on anticipating the dynamics of the interactions between itself and the world.



    Developing a methodology for physics-informed planning requires the careful integration of several capabilities. The robot must perceive and reconstruct its environment with sufficient geometric detail to support nuanced interactions. Yet, a bulk geometric reconstruction alone is not enough; an understanding of the scene's semantics is also required to segment individual entities and assign material characteristics (e.g., density, friction), which directly affect how each object reacts to contact and/or manipulation. Moreover, the system should incorporate a model to accurately predict the outcomes of both robot-object and object-object interactions, enabling the robot to plan actions that reliably anticipate the environment's response.


    Motivated by these challenges and building upon recent breakthroughs in differentiable rendering, foundation models, and physics simulation, we propose \textbf{Scan, Materialize, Simulate} (SMS), a framework for physics-informed robot action planning, comprised of the three eponymous stages:
    \begin{description}[itemsep=0pt, topsep=0pt, partopsep=0pt, leftmargin=0pt]
        \item[\textbf{Scan:}] The robot observes the scene from multiple viewpoints to acquire a set of color-and-depth (RGBD) images used to construct a detailed geometric representation using 3D Gaussian Splatting (3DGS)~\citep{kerbl3Dgaussians}. Visual foundation models (VFMs) perform object detection~\citep{minderer2023scaling} and segmentation~\citep{ravi2024sam2} on the RGBD images, allowing volumetric primitives to be associated with their corresponding objects.
        \item[\textbf{Materialize:}] These volumetric primitives are converted into physics-ready meshes, with estimated material attributes such (e.g., density, friction) inferred by a visual language model (VLM).
        \item[\textbf{Simulate:}] The reconstructed meshes are instantiated in a physics simulation wherein candidate robot actions are optimized before being enacted in the target environment.
    \end{description}
    Collectively, SMS blends the generalist, semantic understanding provided by learning-based foundation models with the principled consistency of physical simulation, enabling generalizable, semantically-aware physics-informed robot planning. We illustrate this framework in \Cref{fig:hero}.

\vspace{-0.1in}
\section{Related Work}
\vspace{-0.1in}
    \textbf{Learning-Based Robot Policies:} Reinforcement learning (RL)~\citep{sutton1999reinforcement, yarats2021image, nair2023r3m} and imitation learning (IL)~\cite{hussein2017imitation, shridhar2023perceiver, florence2022implicit, chi2024diffusionpolicy} have reigned as the dominant paradigms for developing robot policies in recent years. Although physical priors are generally not explicitly enforced, these policies must implicitly encode knowledge of the physical phenomena pertaining to the task at hand (e.g., gravity, gripper friction). This knowledge is provided implicitly via the training environments~\citep{tunyasuvunakool2020, denil2017learning} in the case of RL, or the training demonstrations for IL. However, these learned policies are well known to suffer significant performance degradation when confronted with out-of-distribution (OOD) data~\citep{argall2009survey, sinha2022system}, and we cannot expect their implicit physical knowledge to generalize to the infinitude of possible world configurations a robot may encounter.

    High-capacity vision-language-action (VLA) models~\citep{brohan2023rt, open_x_embodiment_rt_x_2023, kimopenvla, black2024pi0visionlanguageactionflowmodel, wen2025dexvla, huang2025otter} for robotics have also begun to emerge alongside the broader wave of foundation models~\citep{zhou2024comprehensive}. Trained on massive, internet-scale, datasets, these models demonstrate strong generalization, semantic reasoning, and zero-shot performance on a variety of manipulation and navigation tasks. However, their primary strength lies in their semantic and contextual understanding of diverse tasks and scenes, while their demonstrated applications rarely extend to the physically dynamic tasks we consider in this work.

    \textbf{Differentiable Rendering in Robotics:} The advent of modern differentiable rendering methods~\citep{mildenhall2020nerf, fridovich2022plenoxels, mueller2022instant, kerbl3Dgaussians} has provided powerful new tools for scene reconstruction in robotics. Methods built around these representations have quickly made their way into a range of robotics applications, serving as highly detailed obstacle maps for motion planning~\citep{chen2024catnips, chen2024splat, lei2025gaussnav}, providing new means for object tracking~\cite{abou-chakra2024physically, zhangdynamic}, performing system identification and visuomotor control~\citep{gradsim, liudifferentiable}, and underpinning visually accurate virtual environments for policy training in simulation~\citep{low2024sous, li2024robogsim, qureshi2024splatsim, quachgaussian, torne2024reconciling, barcellona2025dream} and synthetic data generation~\citep{meyer2024pegasus}. Semantic features have also been integrated into these reconstructions~\citep{kerr2023lerf, zhou2024feature, qin2024langsplat}, enabling robots to interpret local scene semantics and manipulate objects according to natural language task specifications~\citep{shen2023distilled, lu2024manigaussian}. 
    
    Of particular relevance to this work are the methods which integrate physics into the reconstructed environment geometry, such as~\citep{xie2024physgaussian, qiu-2024-featuresplatting, zhao2025efficientphysicssimulation3d}, which deform the scene geometry according to the material point method (MPM)~\citep{jiang2016material}. However, these methods are primarily oriented towards graphics applications, with the main objective being to produce plausible and visually appealing animated scenery. Some of the aforementioned works such as~\citep{abou-chakra2024physically, torne2024reconciling, zhangdynamic} have explored physically simulating the motion of reconstructed scene components under predicted forces or interactions, but their focus has largely been on manipulation scenarios, often targeting specific object interactions. In contrast, our work aims to enable robots to plan and reason about physically grounded interactions, bridging the gap between physical scene reconstruction and robotic action planning.

    \textbf{Statement of Contributions:} In this work, we present SMS, a novel framework for physics-informed robot action planning. SMS integrates 3D Gaussian Splatting (3DGS) for accurate reconstruction of intricate scene geometry, visual foundation models (VFMs) for semantic understanding and precise object-level segmentation, a vision-language model (VLM) for context-aware material and attribute inference, and a physics engine for consistent prediction of action outcomes based on well-understood Newtonian mechanics. By combining these components, SMS enables generalizable object recognition, scene decomposition, and reliable action-conditioned physical prediction. 
    
    We demonstrate the performance of SMS in a high-precision, billiards-inspired task, where a manipulator must strike a cue ball to direct a target ball to a goal, and a precarious landing scenario, where a quadrotor must safely perch atop an unstable stack of objects. We report performance in simulated domain transfer experiments and validate our method in the real-world with a physical quadrotor. SMS provides a planning strategy that is more generalizable and adaptable than application-specialized algorithms, and delivers superior physical reasoning compared to purely learning-based methods, especially in tasks that demand principled treatment of physical dynamics and interactions.
     
\vspace{-0.1in}
\section{Physics-Informed Planning with SMS}
\vspace{-0.1in}
    In this work, we consider a setting where a robot must perform a task involving physical interaction with the environment, formulated as an optimal control problem. Since modeling the full dynamics of the world is often intractable beyond simple cases, SMS addresses this challenge by reconstructing a surrogate model of the environment. It uses a physics engine to approximate discrete-time dynamics and solves the resulting surrogate optimization problem:
    \begin{equation*}
        \mathrm{minimize}_{\mathbf{u}_{0:T-1}} \quad \hat{L}\big(\hat{\mathbf{x}}_{0:T},\, \mathbf{u}_{0:T-1}\big)
        \quad \text{subject to} \quad
        \begin{aligned}[t]
            & \hat{\mathbf{x}}_{k+1} = \hat{f}(\hat{\mathbf{x}}_k, \mathbf{u}_k), \quad k = 0, \ldots, T-1, \\
            & \hat{\mathbf{x}}_0 = \hat{\mathbf{x}}_\text{init}.
        \end{aligned}
    \end{equation*}
    Here, $\hat{L}$ denotes the surrogate objective, $\hat{f}$ represents the discrete-time surrogate dynamics provided by the physics engine, and $\hat{\mathbf{x}}_k$ is the reconstructed environment's state at time step~$k$. We define the discrete-time state and control trajectories as $\hat{\mathbf{x}}_{0:T} = (\hat{\mathbf{x}}_0, \ldots, \hat{\mathbf{x}}_T)$ and $\mathbf{u}_{0:T-1} = (\mathbf{u}_0, \ldots, \mathbf{u}_{T-1})$, respectively. In the sections that follow, we outline our approach for constructing the virtual environment and solving the associated surrogate optimal control problem.
\vspace{-0.1in}
\subsection{Scan}
\vspace{-0.1in}
    The first step of our pipeline aims to create a detailed 3D reconstruction of the scene. Precise scene geometry is necessary for physical reasoning, but object-level segmentation is essential as it enables the identification and tracking of individual entities, allowing for the simulation of their interactions with each other and with the robot.

    To achieve this, the robot first observes the scene from multiple angles, collecting a sequence of $N$ RGBD frames, $\mathbf{F} = (\mathbf{F}_1, \dots, \mathbf{F}_N)$, along with their associated camera poses. Frame $\mathbf{F}_i$ consists of color $\mathbf{I}_i \in \mathbb{R}^{H \times W \times 3}$ and depth $\mathbf{D}_i \in \mathbb{R}^{H \times W}$, where $H$ and $W$ are the image height and width, respectively. We employ OWLv2~\citep{minderer2023scaling} for zero-shot object detection in the initial image, and use these detections to prompt the Segment Anything Model 2 (SAM 2)~\citep{ravi2024sam2} for segmentation and tracking across the sequence $\mathbf{F}$, resulting in the segmentation frames $\mathbf{S} = (\mathbf{S}_1, \dots, \mathbf{S}_N)$. Each segmentation frame $\mathbf{S}_{i} \in \mathbb{N}^{H \times W}$ specifies an object class $k \in \{1, \dots, K\}$ for each pixel in image $i$.

    We reconstruct the scene using Gaussian splatting~\citep{kerbl3Dgaussians}, which optimizes a collection of 3D Gaussians with indices $\{1,\dots,p\}$ to minimize the difference between rendered and observed images from a set of training viewpoints. A given Gaussian $\mathbf{G}_i$ is parametrized by its position $\mathbf{p}_i \in \mathbb{R}^3$, rotation matrix $\mathbf{R}_i \in \mathbf{SO}(3)$, scale $\textbf{s}_i \in \mathbb{R}^3$, opacity $\mathbf{o}_i \in [0,1]$, and color $\mathbf{c}_i \in \mathbb{R}^3$. Gaussian $i$ can be projected to the pixel coordinates of viewpoint $v$ as $\mathbf{p}_i^\text{image} = \mathbf{M}_v\mathbf{p}_i$, where $\mathbf{M}_v \in \mathbb{R}^{3 \times 4}$ is a projection matrix mapping homogeneous world coordinates to homogeneous pixel coordinates. Given the world-to-camera rotation matrix, $\mathbf{R}_v^{\text{cw}} \in \mathbf{SO}(3)$, the color $C$ at pixel coordinates $\mathbf{u}$ can be rendered from $m$ ordered Gaussians (by increasing distance from the camera) as
    \begin{equation*}
        C(\mathbf{u}) = \sum\nolimits_{j \leq m} \mathbf{c}_j\, \mathbf{\alpha}_j \prod\nolimits_{k < j} (1 - \alpha_k),
    \end{equation*}
    where $\alpha_j = o_j \exp\left(-\frac{1}{2} \left(\mathbf{u} - \mathbf{p}_j^\text{image}\right)^\top \mathbf{\Sigma}_j \left(\mathbf{u} - \mathbf{p}_j^\text{image}\right) \right)$ and $\mathbf{\Sigma}_j = \mathbf{R}_v^{\text{cw}} \operatorname{diag}{(\mathbf{s}_j)^2} \mathbf{R}_v^{\text{cw}\top}$. 
    
    In~\citep{kerbl3Dgaussians}, the parameters of the Gaussians are iteratively optimized via gradient descent to minimize the photometric loss between the rendered and observed image from a sampled training view $\tau$,
    \begin{equation*}
        L_\text{photo}^{(\tau)} = \tfrac{(1 - \lambda_\text{photo})}{HW} \|\hat{\mathbf{I}}_\tau - \mathbf{I}_\tau\|_1 + \lambda_\text{photo} (1 - \text{SSIM}(\hat{\mathbf{I}}_\tau, \mathbf{I}_\tau)),
    \end{equation*}
    where $\lambda_\text{photo}$ is a weighting term, and \text{SSIM} is the structural similarity index measure~\citep{wang2004image}. Hereafter, we use hatted notation to indicate rendered quantities, with the non-hatted counterpart denoting the training datum. Optionally, an isotropic regularization loss can be added to penalize the Gaussians in view from becoming excessively eccentric or skinny:
    \begin{equation*}
        L_\text{isotropic}^{(\tau)} = \tfrac{1}{P}\sum\nolimits_{i=1}^P \mathbf{1}^\top \left|\mathbf{s}_i - \bar{\mathbf{s}}_i\right|,
    \end{equation*}
    where $\mathbf{1}$ indicates a vector of ones and the bar notation indicates the mean value of the vector.

    Beyond just visual appearances, we also need to accurately reconstruct geometric detail. To this end, we respectively render pixelwise depth following~\citep{yugay2023gaussian} and surface normal vectors following~\citep{turkulainen2024dnsplatter} as
    \begin{equation*}
        D(\mathbf{u}) = \sum\nolimits_{j \leq m} \mathbf{p}_j^{\text{image}, z}\, \alpha_j \prod\nolimits_{k < j} (1 - \alpha_k), ~~ \text{and} ~~ \textbf{N}(\mathbf{u}) = \sum\nolimits_{j \leq m} \mathbf{n}_j\, \alpha_j \prod\nolimits_{k < j} (1 - \alpha_k),
    \end{equation*}
    where the normal direction is taken to be a Gaussian's shortest principal axis direction $\mathbf{n}_j = \mathbf{R}_j \operatorname{OneHot}(\arg\min(\mathbf{s}_j))$. Accordingly, we introduce the respective depth and normal loss terms,
    \begin{equation*}
        L_\text{depth}^{(\tau)} = \tfrac{1}{HW}\|\hat{\mathbf{D}}_\tau - \mathbf{D}_\tau\|_1,~~\text{and} ~~
        L_\text{normal}^{(\tau)} = \tfrac{\lambda_\text{normal}}{HW} \sum\nolimits_\mathbf{u}\|\hat{\mathbf{N}}_\tau(\mathbf{u}) - \mathbf{N}_\tau(\mathbf{u})\|_1 + \tfrac{\lambda_\text{scale}}{P}\sum\nolimits_{i=1}^P \|\arg\min(\mathbf{s}_i)\|_1.
    \end{equation*}
    Depth maps are directly obtained from RGBD data, while normal vectors can be computed from the observed depth analytically~\citep{zhouopen3d}. Note we add an extra term in $L_\text{normal}^{(\tau)}$ to promote the optimization of flat, disk-like Gaussians to more closely map the scene surface geometry. We control the normal mapping and Gaussian scaling regularization with $\lambda_\text{normal}$ and $\lambda_\text{scale}$, respectively.
    
    Finally, to enable entity-wise segmentation, we augment the Gaussians with randomly initialized affinity feature vectors $\mathbf{a}_i \in \mathbb{R}^d$, which are optimized to associate each Gaussian with one of the $K$ entities in the scene. We render these features as 
    \begin{equation*}
        \mathbf{A}(\mathbf{u}) = \sum\nolimits_{j \leq m} \mathbf{a}_j\, \alpha_j \prod\nolimits_{k < j} (1 - \alpha_k),
    \end{equation*}
    and pass them through a classifier $\phi$ (implemented as a two-layer neural network, trained jointly with the Gaussians) to obtain logits $\mathbf{p}(\mathbf{u}) = \phi(\mathbf{A}(\mathbf{u})) \in \mathbb{R}^K$ and predicted object label at pixel $\mathbf{u}$ as $y(\mathbf{u}) = \operatorname{softargmax}(\mathbf{p}(\mathbf{u}))$.
    
    To optimize the affinity features and the classifier, we use the cross-entropy loss between the predicted logits $p_c$, for each class $c$, and the assigned label $\mathbf{S}(\mathbf{u})$ for each pixel $\mathbf{u}$:
    \begin{equation*}
        L_\text{segmentation}^{(\tau)} = -\tfrac{1}{HW}\sum\nolimits_{\mathbf{u}} \log \left( 
\exp\left(p_{\mathbf{S}_\tau(\mathbf{u})}(\mathbf{u}) \right)/\sum\nolimits_{c=1}^K \exp \left( p_c (\mathbf{u})\right)\right).
    \end{equation*}
     Similar to \citep{yugay2023gaussian}, our implementation of Gaussian splatting optimizes each frame sequentially. Upon reaching a new frame, Gaussians are seeded from a subset of the back-projected RGBD point cloud, with sampling biased towards low-density, unmapped regions, or regions with high depth error. At the first frame, points are sampled primarily from edges detected from the color image, we perform 1000 gradient descent steps. Subsequent frames are optimized for 100 steps. While optimizing a particular frame, a gradient descent step is occasionally taken for a randomly sampled previous frame to avoid catastrophic forgetting. The overall loss that we compute for frame viewpoint $\tau$ is:
    \begin{equation*}
        L^{(\tau)} = L_\text{color}^{(\tau)} + L_\text{isotropic}^{(\tau)} + L_\text{depth}^{(\tau)} + L_\text{normal}^{(\tau)} + L_\text{segmentation}^{(\tau)}.
    \end{equation*}

\vspace{-0.1in}
\subsection{Materialize}
\vspace{-0.1in}
    Once the environment is mapped and distinct entities have been localized, we must convert the Gaussian splatting reconstruction into a set of simulation-ready meshes, a format compatible with most physics simulators~\citep{todorov2012mujoco, coumans2020, Genesis}. Entity geometries are recovered by grouping Gaussians using their affinity features, with their centers forming an initial object point cloud. Since the optimized Gaussians can vary significantly in scale and orientation, this point cloud often yields nonuniform surface coverage. As such, we exploit each Gaussian's extended geometry which was optimized to be flat, disk-like, and tightly adherent to the scene surface by locally sampling additional points from the plane defined by its two largest principal axes. This produces a uniformly dense surface point cloud for subsequent processing.
    
    For irregularly shaped objects, we perform a Delaunay tetrahedralization~\citep{delaunay1934sphère, barber1996quickhull} on the densified point cloud to generate a volumetric mesh. This tetrahedralization yields the convex hull of an object, but we can recover non-convex geometry by pruning tetrahedra with edges exceeding a set, or adaptively determined, length threshold. The final object mesh is extracted by retaining only the external faces and vertices of this pruned tetrahedralization. For entities identified as approximately spherical (such as balls), we instead fit a sphere to the densified point cloud via random sample consensus (RANSAC)~\citep{fischler1981random} in order to preserve rolling dynamics in simulation.
    
    Finally, to support realistic physical behavior, we prompt OpenAI's GPT-4o VLM with an image of each detected entity, querying it for the dominant material class along with relevant properties such as the density and the coefficients of friction and restitution. Leveraging a VLM as such enables recognition of arbitrary, previously unseen objects and taps into the broad material knowledge encoded in the large language model backbone. This approach offers far greater flexibility than manually maintained lookup tables or fixed classifiers, allowing our system to infer physical properties for a wide range of objects encountered in real-world environments with minimal prior information.

\vspace{-0.1in}
\subsection{Simulate}
\vspace{-0.1in}
    The meshed entities are imported into a physics simulator together with a detailed model of the robot, which is readily available in a compatible format for most popular platforms. We define a task-specific objective function and optimize the robot actions using the simulator as a rigid-body dynamics constraint module, predicting the environment’s response to candidate actions.

    Given the highly nonlinear and potentially discontinuous nature of these dynamic tasks, we employ the Nelder-Mead simplex optimization algorithm~\citep{nelder1965simplex}, which is a robust, gradient-free method well suited for low- to moderate-dimensional problems. The optimizer iteratively refines candidate actions until convergence or a fixed iteration budget is reached, after which the best solution found can be deployed in the target environment.
    
\vspace{-0.1in}
\section{Experiments}
\vspace{-0.1in}
    To assess the performance of SMS, we conduct experiments on two challenging domains: a billiards-inspired manipulation task and a quadrotor landing scenario. In both settings, we evaluate SMS using simulated domain transfer experiments with the target environment constructed in Nvidia Isaac Sim, a photorealistic platform with high-fidelity physical simulation. We also demonstrate a landing scenario with a physical quadrotor to validate our method's real-world applicability. We briefly explain the experimental setup in these sections, but provide specific details in the appendix.

\vspace{-0.1in}
\subsection{Billiards Scenario}
\vspace{-0.1in}
    In the billiards-inspired task, a Franka Research 3 (FR3) robot is positioned on a flat surface. In each scene, a cue ball is placed in front of the robot within a defined region, while a target billiard ball and a goal position are randomly sampled elsewhere in the scene. Several obstacles representing common household objects (e.g., toys, dishware, decor) of varied geometry and material are randomly selected from the photorealistic asset library in Nvidia Omniverse and scattered throughout the workspace. Entities in the scene, including the robot and floor, are manually assigned realistic physical parameters, such as friction, restitution, and mass, to ensure realistic behavior. We reconstruct each scene using 60 images, with optimization averaging 3 seconds per frame for a total mapping time of around 3 minutes.

    \begin{figure}[ht]
        \centering
        \includegraphics[height=5.65cm]{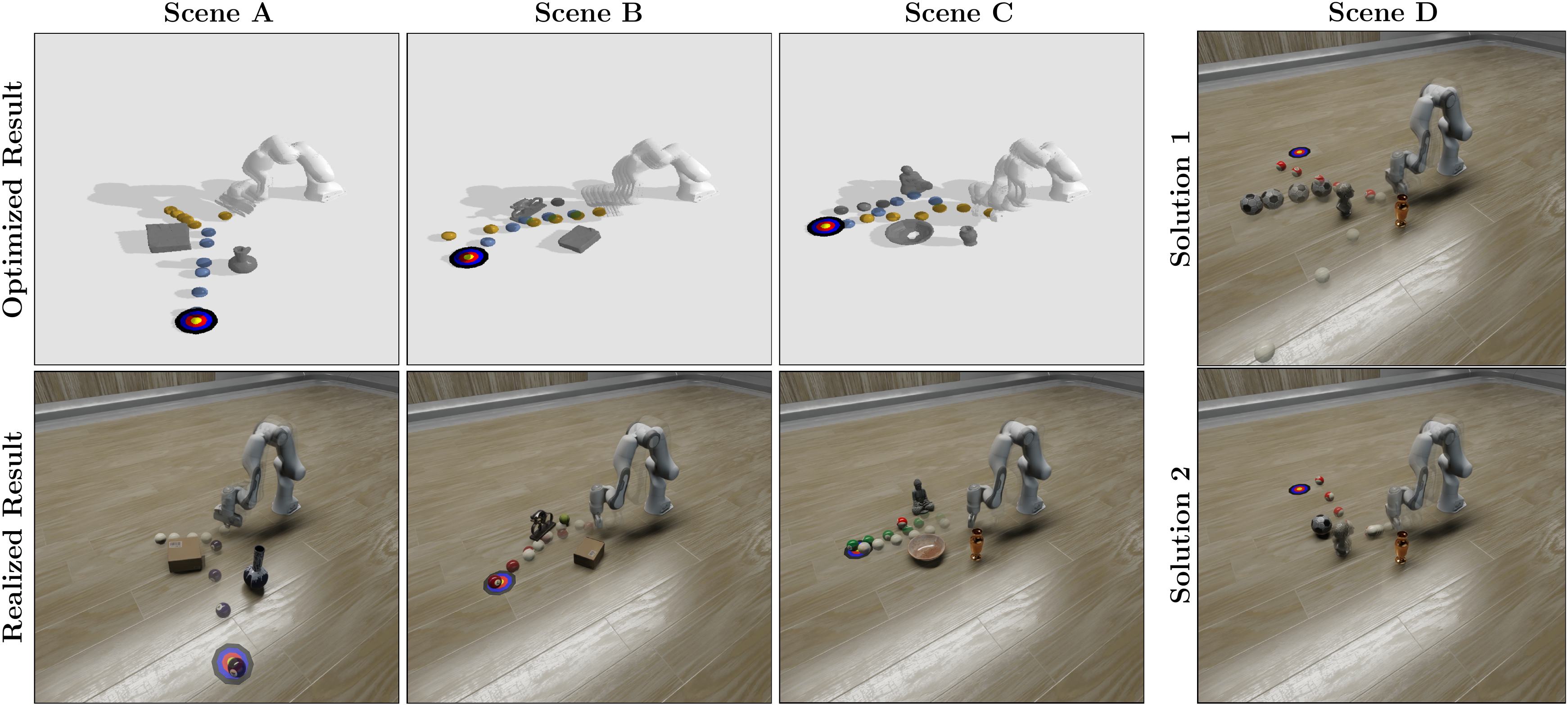}
        \caption{Billiards scenes. Left: Motion composite images show the actions optimized in the reconstructed virtual environment (top) and subsequently realized in the target environment (bottom) for Scenes A, B, and C (cue and target balls colored yellow and blue, respectively, and the target is indicated by the bullseye pattern). Scene A depicts a direct shot from cue to target ball. Scene B involves a rebound off the heavy brass sculpture. Scene C features a complex double-rebound involving the Buddha statue and the billiard ball. Right: For the more challenging Scene D, SMS finds two distinct solutions, utilizing rebounds off the soccer ball (top) and the marble statue (bottom).}
        \label{fig:combined_grids}
    \end{figure}


    The robot's task is to strike the cue ball with its end effector fingers to drive the target ball as close as possible to the specified goal. The robot's trajectory, $\mathbf{u}_{0:T-1}$, is parametrized by a contact position $\mathbf{p}_\text{contact}$, a contact speed $v_\text{strike}$, and strike angle $\theta_\text{strike}$. The robot executes a linear sweeping motion, accelerating to reach the contact position (i.e., the estimated cue ball center position) with the desired velocity. The strike angle determines the orientation of this linear path about the contact point. The task objective is defined as the minimum Euclidean distance between the target ball and the goal over the planning horizon, i.e., $\hat{L}_\text{bill} = \min_{t \in \{0,1,\dots,T\}}\|\mathbf{p}_{\text{ball},t} - \mathbf{p}_{\text{goal},t}\|_2$, encouraging the target ball to pass over the goal position at some point during execution, analogous to sinking a ball in pool or putting in golf. Actions are optimized using PyBullet~\citep{coumans2020} as our virtual physics environment. Nelder-Mead optimization averages 1.1 seconds per iteration and typically converges within 15 to 20 of the allotted 30 iterations.

\begin{table}[!htbp]
    \setlength{\abovecaptionskip}{2pt}  
    \setlength{\belowcaptionskip}{2pt}  
    \setlength{\textfloatsep}{5pt}      
    \centering
    \caption{%
        Billiards performance for SMS and the baseline. Values are mean (standard deviation) of the task objective $\hat{L}_\text{bill}$ in centimeters. ``Predicted" and ``realized" refer to performance in the virtual and target environments, respectively.
    }
    \renewcommand{\arraystretch}{0.95}  
    \setlength{\tabcolsep}{3pt}
    \small
    \begin{tabular}{lcccc}
        \toprule
        \rowcolor{white}
        \textbf{Performance} & \textbf{SMS Predicted} & \textbf{SMS Realized} & \textbf{Baseline Predicted} & \textbf{Baseline Realized} \\
        \midrule
        \rowcolor{gray!20}
        \textbf{Overall} & 5.5 (12.7)\textsuperscript{a} & 20.5 (17.9)\textsuperscript{a} &  – &  – \\
        \textbf{Per-Scene Best} & 0.0 (0.0)\textsuperscript{b} & 5.5 (8.3)\textsuperscript{b} & 5.76 (10.2)\textsuperscript{b} & 28.2 (21.7)\textsuperscript{b} \\
        \rowcolor{gray!20}
        \textbf{Per-Scene Worst} & 27.1 (19.3)\textsuperscript{b} & 41.2 (22.1)\textsuperscript{b} & – & – \\
        \bottomrule
    \end{tabular}
    \vspace{-0.3em}  
    \parbox{\linewidth}{%
        \footnotesize
        \textsuperscript{a}~Aggregated over all 18 scenes $\times$ 30 seeds for SMS; not applicable to deterministic baseline.\\
        \textsuperscript{b}~Mean and standard deviation across 18 scenes (best or worst per-scene).
    }
    \vspace{-1em}  
    \label{tab:manipulator_results}
\end{table}


    We evaluate SMS over 18 generated scenes. Since our optimizer is sensitive to the initial guess, we optimize each scene from 30 random initializations. We present quantitative performance figures in \Cref{tab:manipulator_results} aggregated across all scenes. The table compares the task objective function predicted for the optimal result in the virtual environment and the realized objective in the target environment. We report the mean and standard deviation of the objective function across all evaluations, as well as those across the individual best and worst evaluation in each scene.
    
    In \Cref{tab:manipulator_results}, we also include results for our baseline comparison method. This baseline searches over a dense grid of candidate contact speeds and strike angles at the cue ball position, assuming the initial cue-target ball collision is elastic. To account for future collisions with obstacles, a top-down RGBD image is used to construct a static obstacle mask and trace the post-collision path of the target ball with impacts reflecting the target ball's normal velocity component. Since the baseline produces a single deterministic action, the performance is computed over the 18 per-scene trials and compared against SMS's mean per-scene best performance in \Cref{tab:manipulator_results}. We find that our method's best performing action per scene consistently outperforms this baseline. The baseline can be competitive in simple scenes with few collisions or heavier, immobile obstacles. However, SMS especially outperforms this baseline in scenes with complex or dynamic interactions.

    \begin{wrapfigure}{r}{0.525\linewidth}
        \vspace{-4pt} 
        \includegraphics[width=\linewidth]{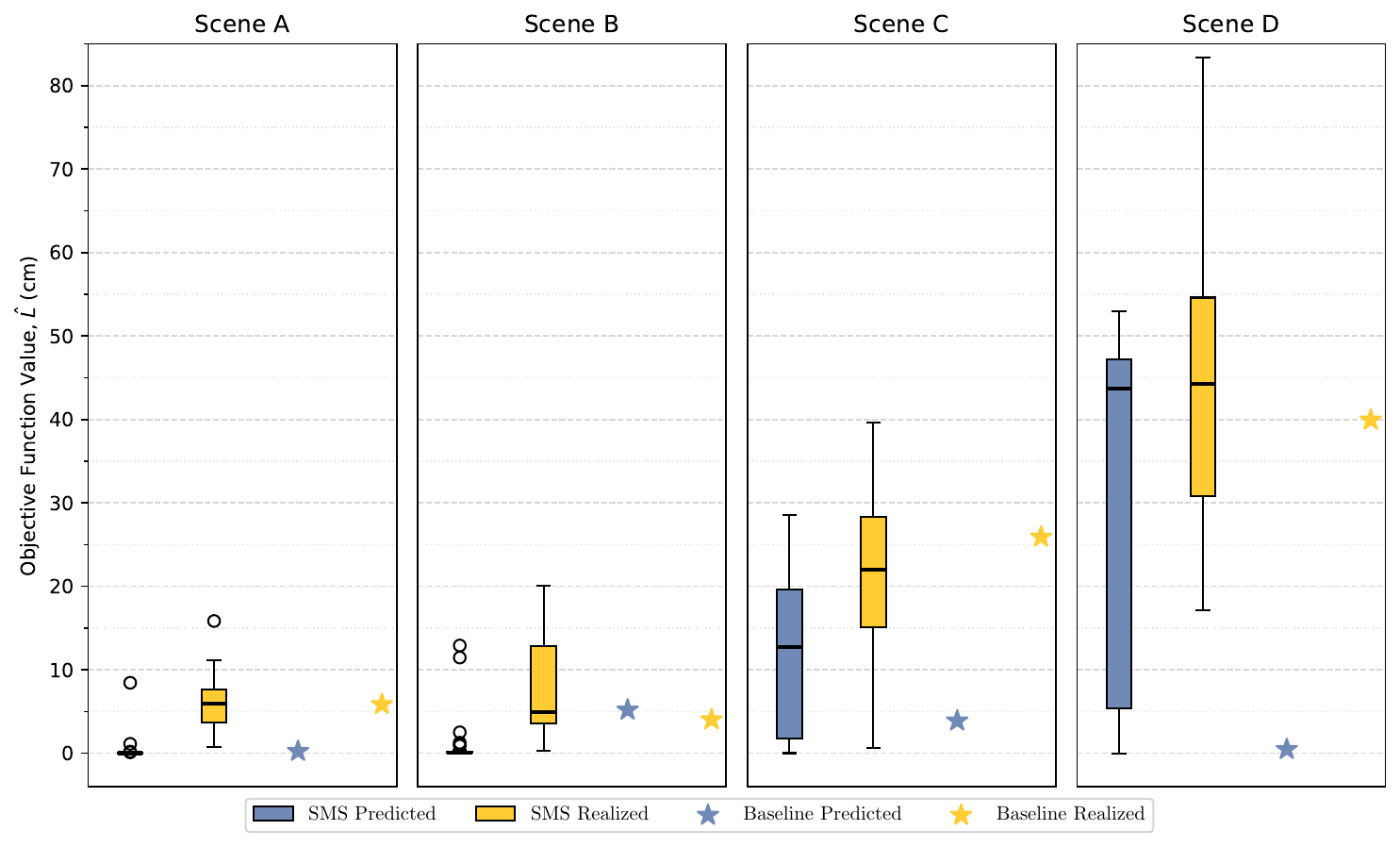}
        \caption{Distributions of SMS performance over 30 repeated action optimizations. Baseline results are shown for comparison. Lower is better.}
        \label{fig:experiment_manipulator_dists}
        \vspace{-4pt} 
    \end{wrapfigure}
    

    To illustrate SMS in action, we showcase four diverse scenes in \Cref{fig:combined_grids}, each varying in object material and geometry. These visualizations demonstrate not only the range of environments considered but also the nuanced behavior enabled by our planning approach. In Scene A, SMS selects a direct, traditional shot from cue to target ball. In Scene B, the planner exploits physical properties, opting to rebound the target ball off the heavy brass sculpture (whereas in Scene A, the lightweight cardboard box is avoided as an ineffective intermediary). Scene C exhibits even greater complexity: the solution involves an initial sharp-angle rebound off the Buddha statue, followed by a second corrective collision with the billiard ball to reach the goal - a result that could not be produced by our baseline. Notably, the actions optimized in the virtual environment translate well to execution, with predicted outcomes closely matching realized trajectories across all scenes. Finally, Scene D illustrates the method’s ability to discover multiple valid strategies, utilizing rebounds off the marble statue and soccer ball, once again demonstrating awareness of how object geometry and material affect momentum transfer.

    We further compare the distributions of predicted and realized performance for these scenes in \Cref{fig:experiment_manipulator_dists}. Some degradation in realized performance is observed, with the difference primarily attributable to domain mismatch, mesh reconstruction errors, and discrepancies in material property estimates. The impact is most pronounced in scenes with high object density, such as Scene C, where small errors compound through successive collisions, and in long-range shots like Scene D, which amplify trajectory divergence. Nonetheless, these experiments highlight SMS's ability to account for scene geometry and material properties, while frequently uncovering multiple effective strategies for task completion.
    



\vspace{-0.1in}
\subsection{Quadrotor Landing Scenario}
\vspace{-0.1in}
    In this setting, a quadrotor is tasked with safely landing on an unstable landing structure constructed by placing a flat landing platform (e.g., a cardboard box or plastic tote) atop a supporting base such as a step ladder, stool, or recycling bin. The landing platform is positioned in such a way that it signficantly overhangs the base, making landing site selection critical to avoid causing the platform or structure to topple. As in the billiards scenario, objects are drawn from Nvidia's Omniverse asset library and assigned realistic physical properties, and each scene is reconstructed using 60 images.

    Beyond the quadrotor's landing impulse, the propeller wash can also impart significant forces on the landing structure on approach. Specifically, flying over an overhanging region of the landing platform can induce moment about the base, causing it to topple before the quadrotor has a chance to land. As such, for this task we jointly optimize the quadrotor's approach path and landing position. We parametrize the quadrotor's approach path by the quadrotor's (fixed) initial position $\mathbf{p}_{\text{quad},0}$, landing position $\mathbf{p}_\text{land}$, and an approach radius $r_\text{app}$ and angle $\theta_\text{app}$. The latter two parameters define the approach waypoint $\mathbf{p}_\text{app}$ relative to the landing position. The approach path is constructed as a B\'ezier curve with control point sequence $(\mathbf{p}_{\text{quad},0}, \mathbf{p}_\text{app}, \mathbf{p}_\text{land})$. We define the the task objective as $\hat{L}_{\text{land}} = \sum_{k=1}^K \| \mathbf{p}_{k,T} - \mathbf{p}_{k, 0} \|_2 + \sum_{k=1}^K 2 \arccos\left(|\mathbf{q}_{k,T} \cdot \mathbf{q}_{k,0}|\right) + \|\mathbf{p}_\text{quad, T} - \mathbf{p}_\text{land}\|_2$, where $\mathbf{p}$ indicates a position, $\mathbf{q}$ is an orientation represented as a quaternion, $k$ indexes an object, and $T$ is the simulated horizon. The first two terms penalize any position and rotation changes of the landing structure, while the last encourages landing position stability.

    \begin{wrapfigure}{r}{0.65\linewidth}
        \vspace{-4pt} 
        \includegraphics[width=\linewidth]{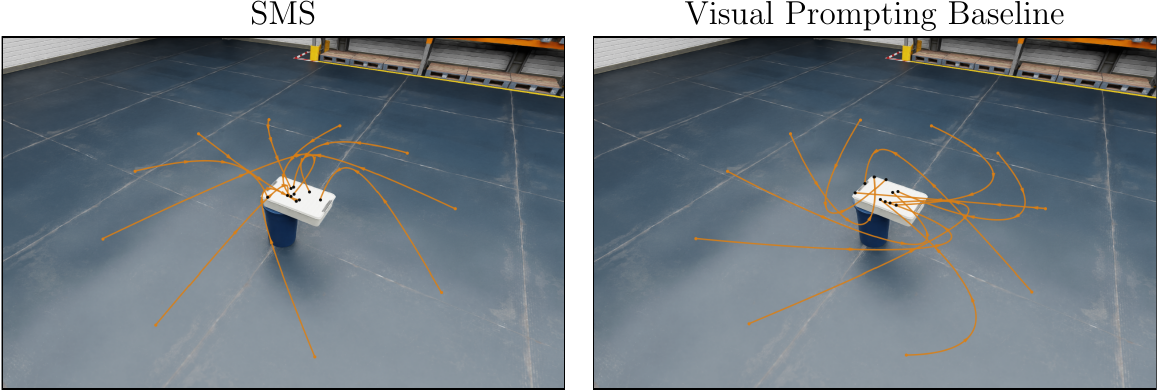}
        \caption{Quadrotor approach comparison. Left: SMS optimizes approach paths that avoid the overhanging ledge. Right: Visual prompting does not consider propeller wash, resulting in flight over the overhang. Note that several landing sites are chosen at the edge and corner of the landing platform.}
        \label{fig:approach_comparison}
        \vspace{-4pt} 
    \end{wrapfigure}
    
    Genesis~\citep{Genesis} is used as the virtual physics environment and scale a provided quadrotor model to match reference real-world dimensions and mass. During simulation, the quadrotor travels to the landing position following the approach B\'ezier and cuts off the motors upon arrival. Propeller wash is modeled using smoothed-particle hydrodynamics (SPH)~\citep{koschier2020smoothed}, with each propeller represented as a discrete fluid emitter with parameters estimated via a simple momentum conservation model. Running Genesis on an Nvidia RTX 4090 GPU, each Nelder-Mead optimization averages 8.2 seconds per iteration for a maximum of 30 iterations.


    We evaluate SMS on four hand-designed scenes, shown in the appendix. For each scene, approach paths and landing positions are computed for the quadrotor starting from 10 uniformly spaced positions on a 2-meter radius circle around the landing platform. As a baseline, inspired by~\citep{nasiriany2024pivot}, we annotate candidate landing positions on an image observed by the quadrotor and prompt GPT-4o to select the best one. Finally, we annotate several candidate paths to the chosen landing site, again querying GPT-4o to select the preferred approach. Across the four scenes, SMS outperforms this baseline with respective landing success rates of 100\%, 80\%, 90\%, and 90\%, compared to 50\%, 50\%, 60\%, and 50\% for the visual prompting strategy.

    \begin{wrapfigure}{l}{0.65\linewidth}
        \vspace{-4pt} 
        \includegraphics[width=\linewidth]{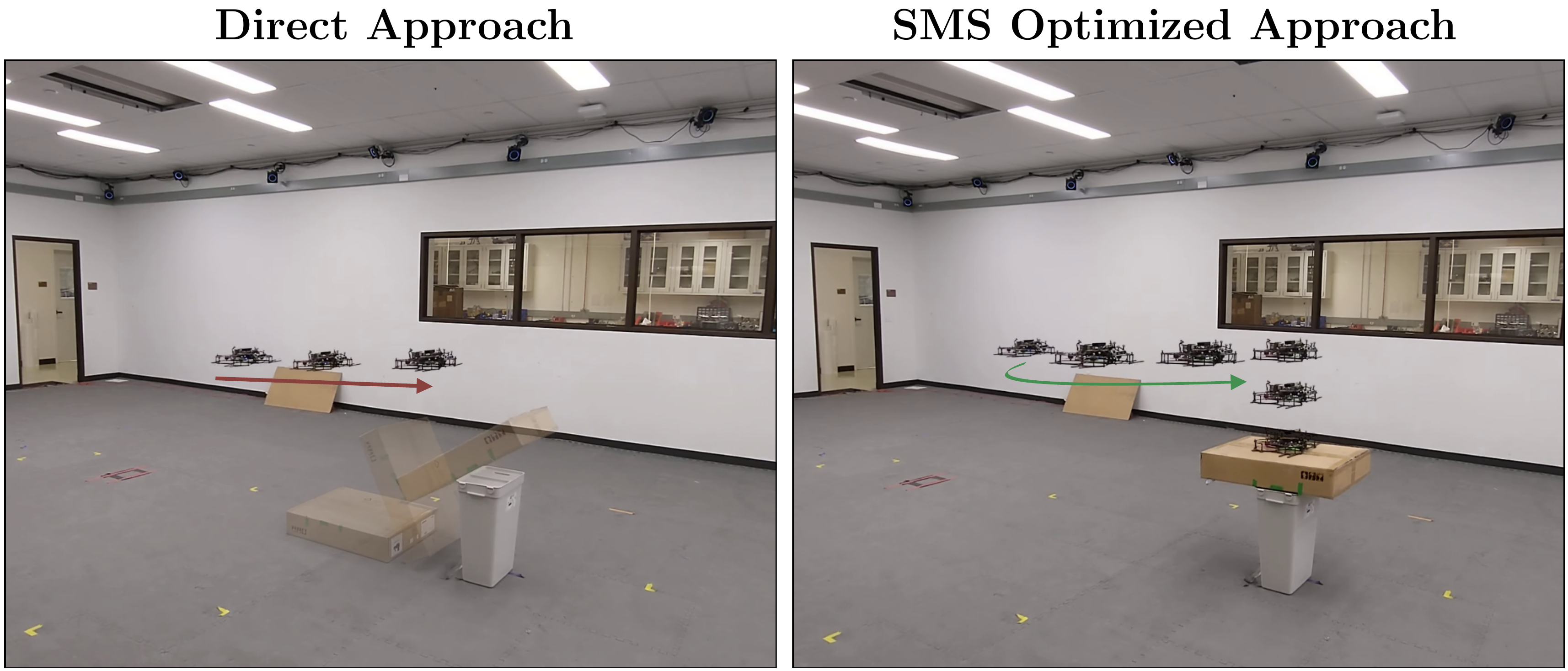}
        \caption{Quadrotor hardware demonstration. Left: A direct approach causes the propeller wash to topple the landing platform. Right: SMS optimizes a roundabout trajectory that avoids disturbing the box and enables a successful landing.}
        \label{fig:quadrotor_hardware}
        \vspace{-4pt} 
    \end{wrapfigure}

    These experiments underscore that, while a VLM provides strong semantic and visual understanding of the scene, it alone does not yield a reliable policy for safety-critical tasks like quadrotor landing. In contrast, SMS grounds the planning in physical simulation, enabling the identification of both stable landing sites and comparatively lower-risk approach trajectories as shown in \Cref{fig:approach_comparison}. This integration results in consistently higher success rates and robust performance, especially in scenarios where the purely visual reasoning falls short, such as when dealing with complex effects like propeller wash. Real-world quadrotor flight tests (\Cref{fig:quadrotor_hardware}) further demonstrate the practical value of physics-informed planning for safe and reliable operation.

    
    
\vspace{-0.1in}
\section{Conclusion, Limitations, and Future Work}
\vspace{-0.1in}

    In this work, we introduced SMS, a generalizable framework for physically grounded robot planning. Our method generates interactive, object-aware scene models to optimize robot behaviors that produce complex, physically dynamic interactions—a capability we demonstrated in both billiards-inspired manipulation and quadrotor landing tasks. By combining the semantic knowledge and generalization ability of foundation models with the principled predictive power of physical simulation, SMS addresses the limitations of prior approaches: specialized methods often fail to generalize beyond narrow domains; learning-based methods struggle to account for novel or out-of-distribution physical behaviors; and approaches relying solely on foundation models sacrifice the precision required for tasks demanding exact physical outcomes. 
    
    In future work, we aim to expand these physical reasoning capabilities to a wider range of robots and environments, and identify several key directions for further development.


    \textbf{Towards Improved Scene Reconstruction:} While Gaussian splatting provides high-quality reconstructions, our implementation requires 2 to 3.5 seconds per frame; with our experiments using 60 frames per scene, reconstruction takes several minutes. Moreover, as with all differentiable rendering methods, unobservable or heavily occluded regions cannot be accurately reconstructed. This proved especially challenging for the interfaces between stacked objects in our quadrotor experiments, which necessitated post-processing to ensure the interface geometries were reasonably modeled. Incorporating generative models for 3D scene completion, such as~\citep{fan2024instantsplat, yangstorm, szymanowicz2025bolt3d}, could simultaneously address both challenges: reducing reconstruction time and leveraging learned priors to plausibly infer occluded or unseen regions.

    \textbf{Incorporating Feedback for Closed-Loop Planning:} A current limitation of SMS is its open-loop design: we query material properties, optimize actions in simulation, and execute them without incorporating feedback. As a result, SMS cannot adapt to inaccurate attribute estimates or unforeseen changes during execution. Incorporating strategies for visual feedback and system identification~\citep[e.g.,][]{gradsim, le2023differentiable, abou-chakra2024physically} could address these limitations by enabling online updates to the scene reconstruction and physical parameters based on observed outcomes.

    \textbf{Expanding Action Expressivity:} In our implementation, SMS uses a gradient-free optimization strategy, which is effective for action spaces that can be compactly parameterized. However, scaling to more expressive or high-dimensional control behaviors would benefit from differentiable physics simulators that support efficient gradient-based optimization.\footnote{At present, no differentiable physics simulator met our project requirements. While we attempted to use the differentiable capabilities of Nvidia Warp~\citep{warp2022}, we encountered numerical instability without prohibitively small time steps. Genesis~\citep{Genesis} has announced upcoming support for differentiable rigid-body simulation, but this was not available at the time of writing.} Furthermore, language-based task specification~\citep{huang2023diffvl} can facilitate objective function formulation for more complex applications and extend the utility of this framework to non-expert users.

\clearpage
\acknowledgments{This work was funded by the DARPA TIAMAT program.}


\bibliography{references}  

\clearpage
\newpage

\appendix

\section{Additional Details for Billiards Scenario}

\renewcommand{\thefigure}{A.\arabic{figure}}
\setcounter{figure}{0}

\vspace{-0.1in}
\subsection{Scene Generation}
\begin{figure}[htbp]
        \newcommand{\imgwidth}{0.33\linewidth}
        \centering
        \setlength{\tabcolsep}{0.5pt}
        \renewcommand{\arraystretch}{0.0}
        \begin{tabular}{ccc}
            \includegraphics[width=\imgwidth]{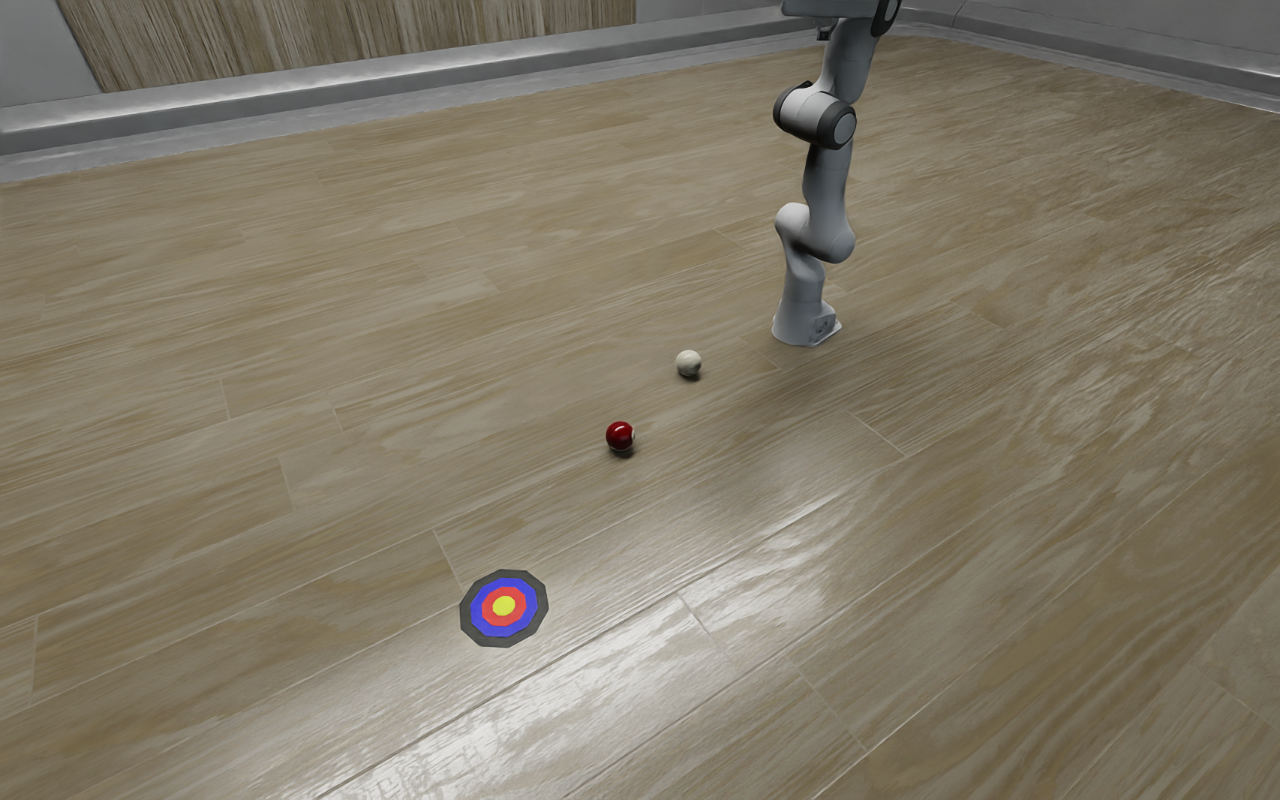} &
            \includegraphics[width=\imgwidth]{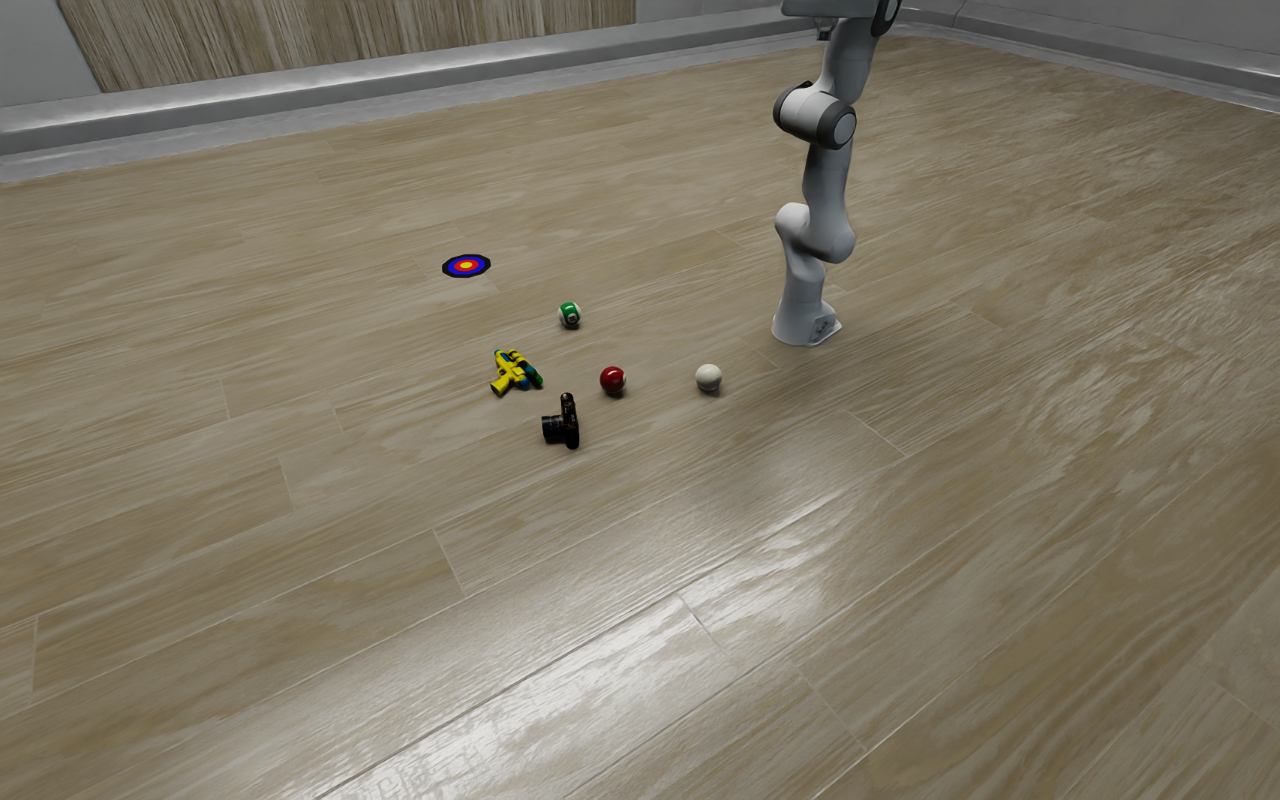} &
            \includegraphics[width=\imgwidth]{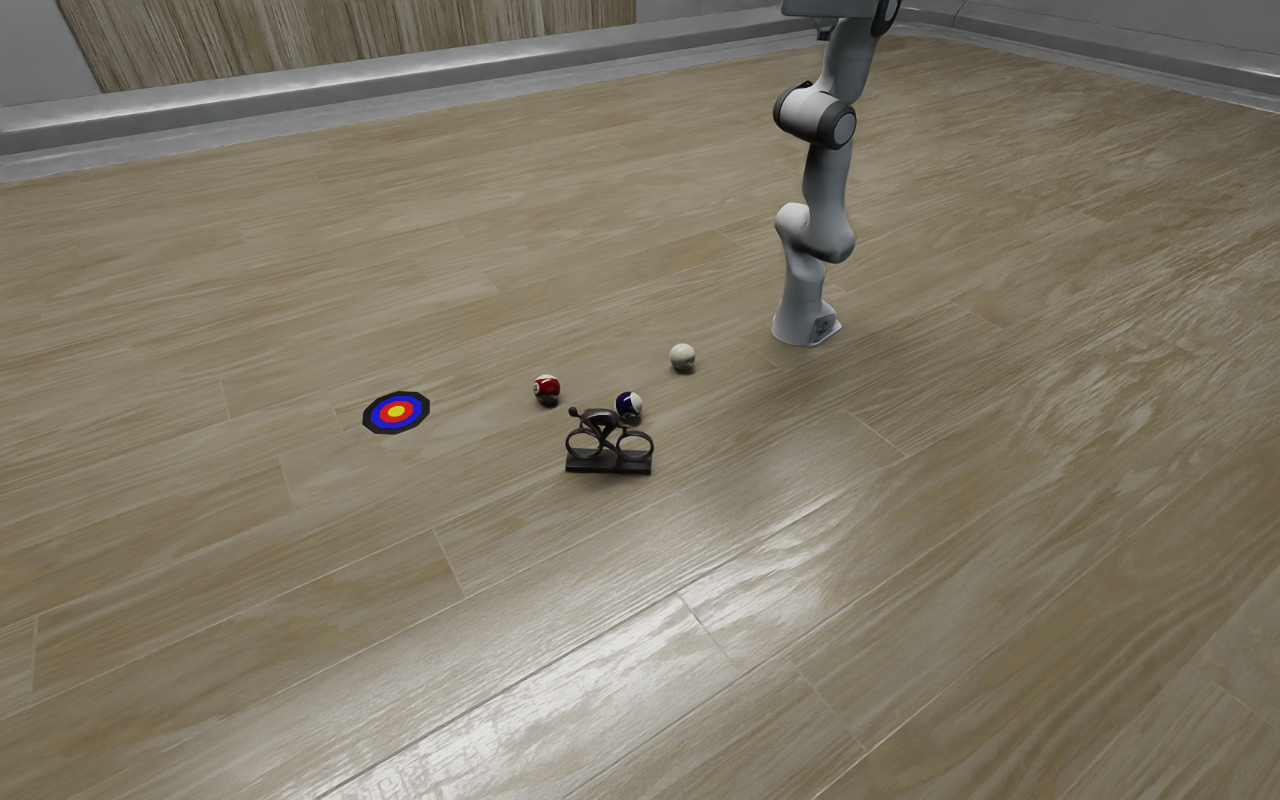} \\
            \noalign{\vskip 1pt}
            \includegraphics[width=\imgwidth]{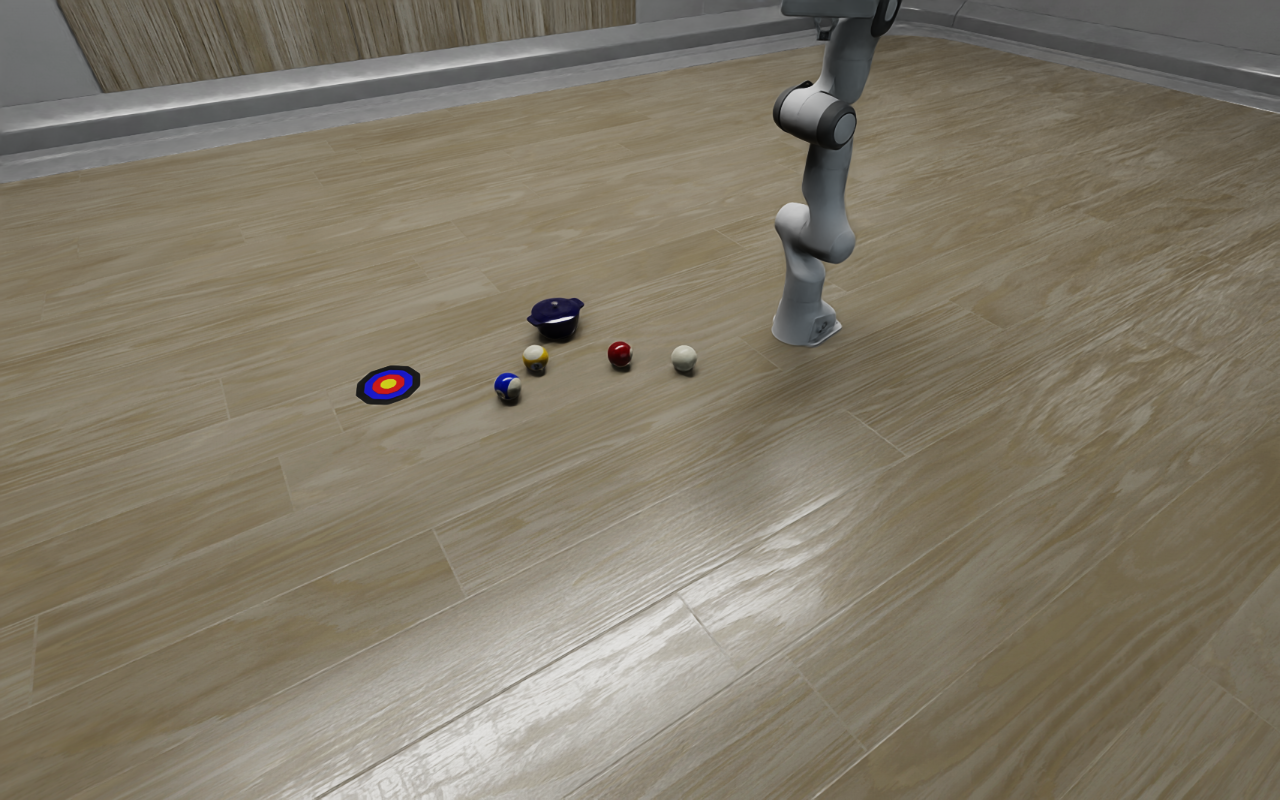} &
            \includegraphics[width=\imgwidth]{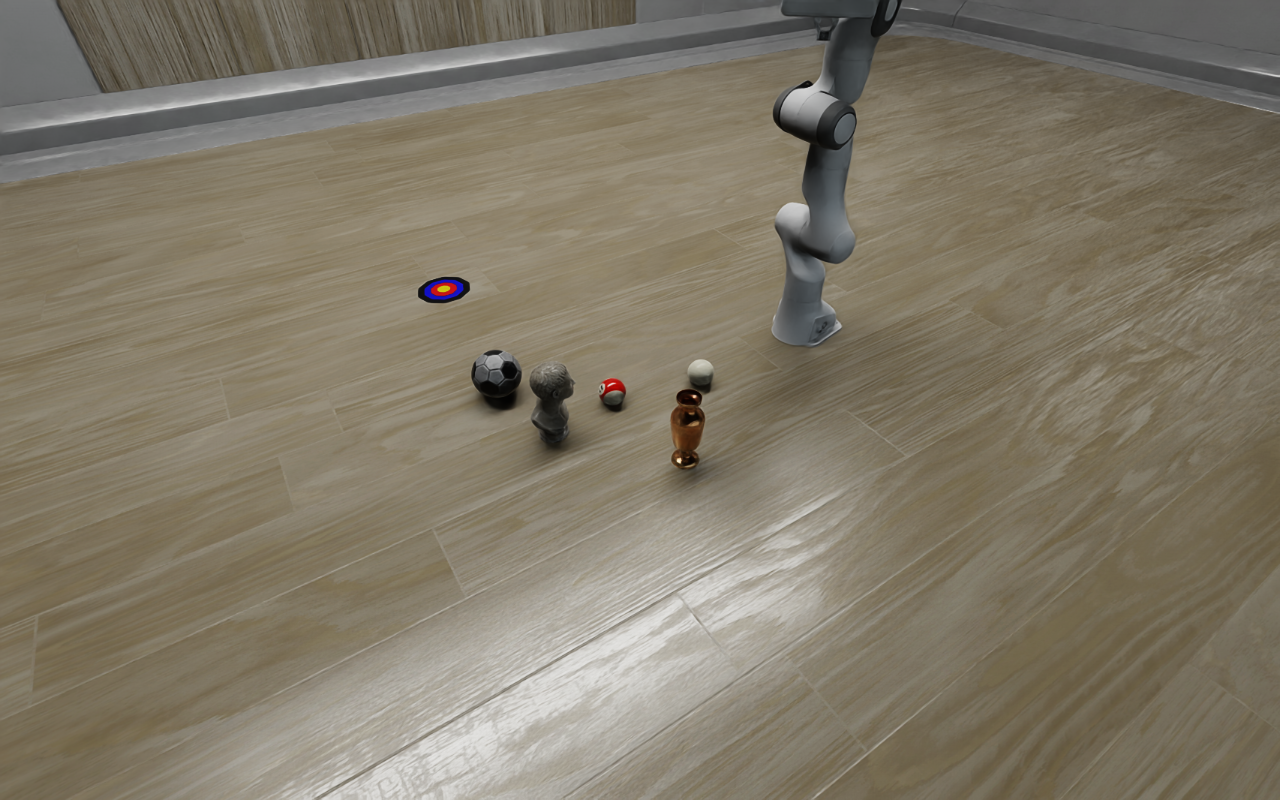} &
            \includegraphics[width=\imgwidth]{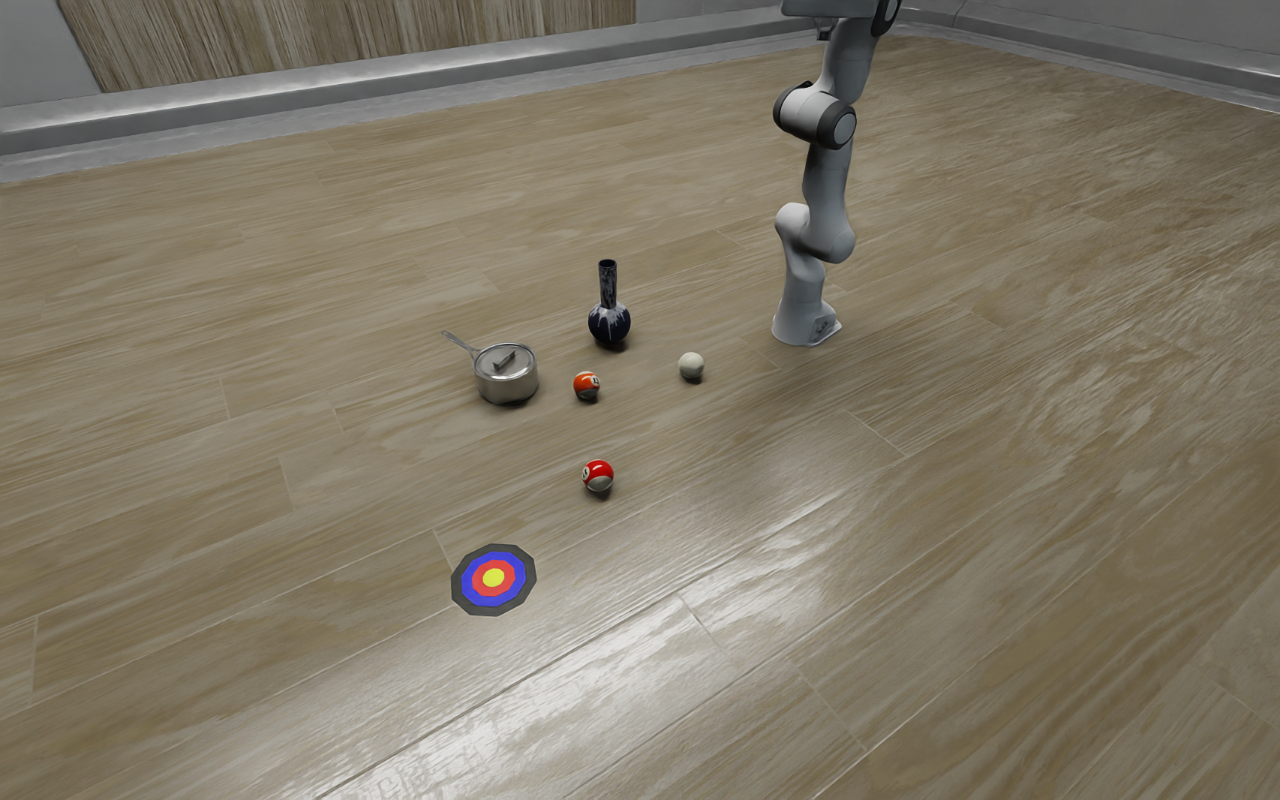} \\
            \noalign{\vskip 1pt}
            \includegraphics[width=\imgwidth]{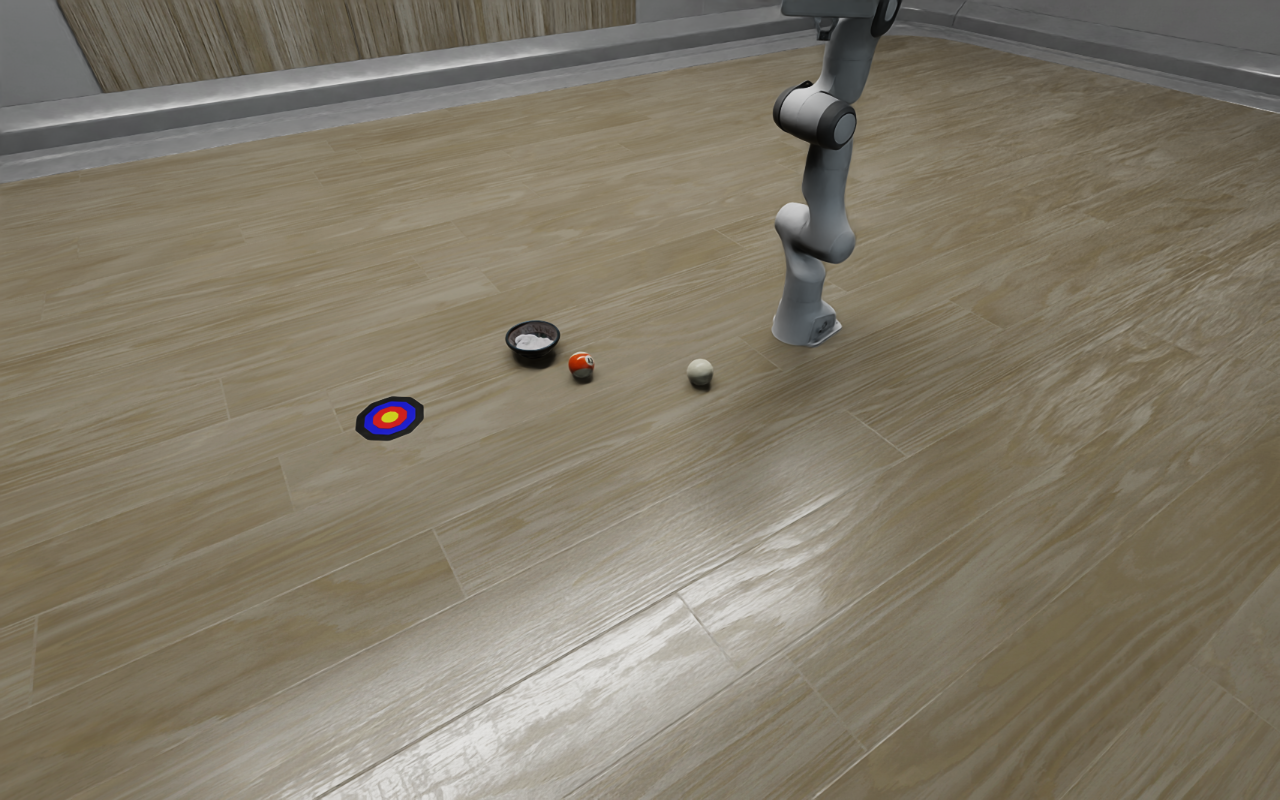} &
            \includegraphics[width=\imgwidth]{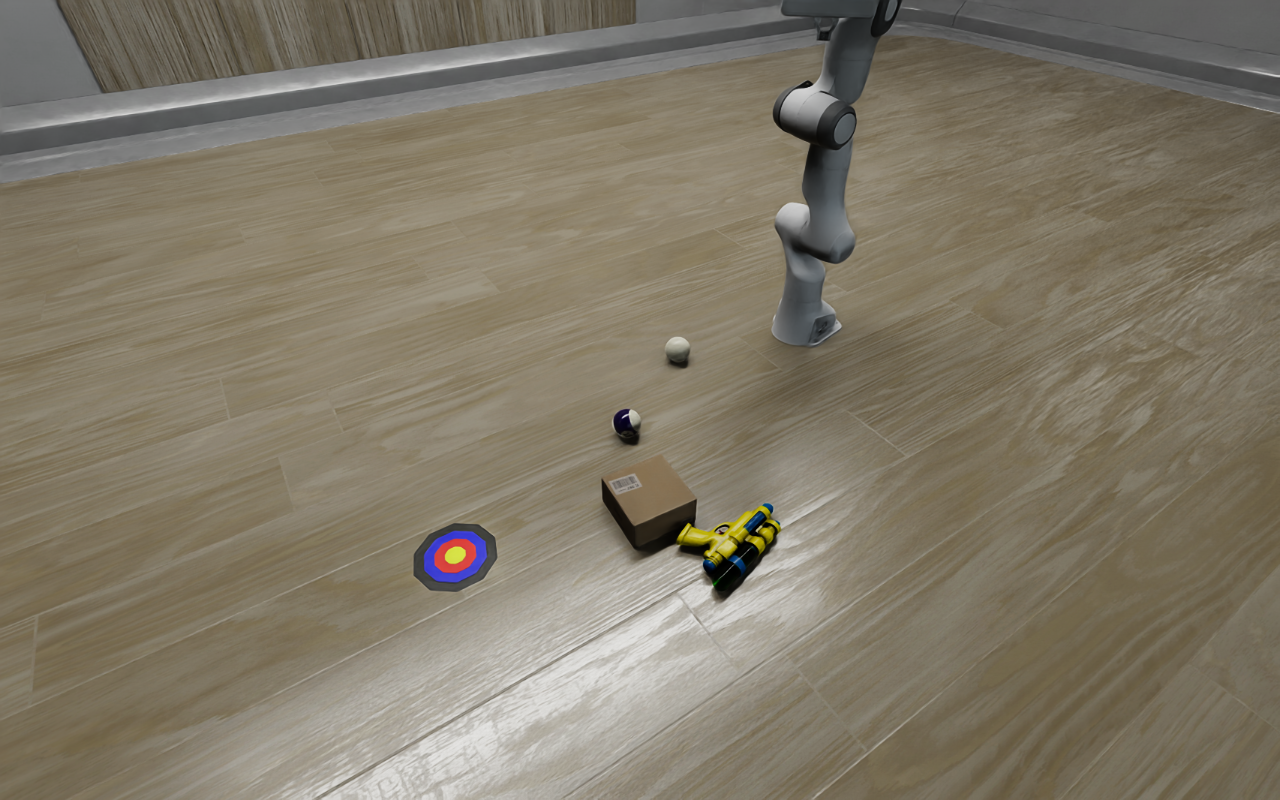} &
            \includegraphics[width=\imgwidth]{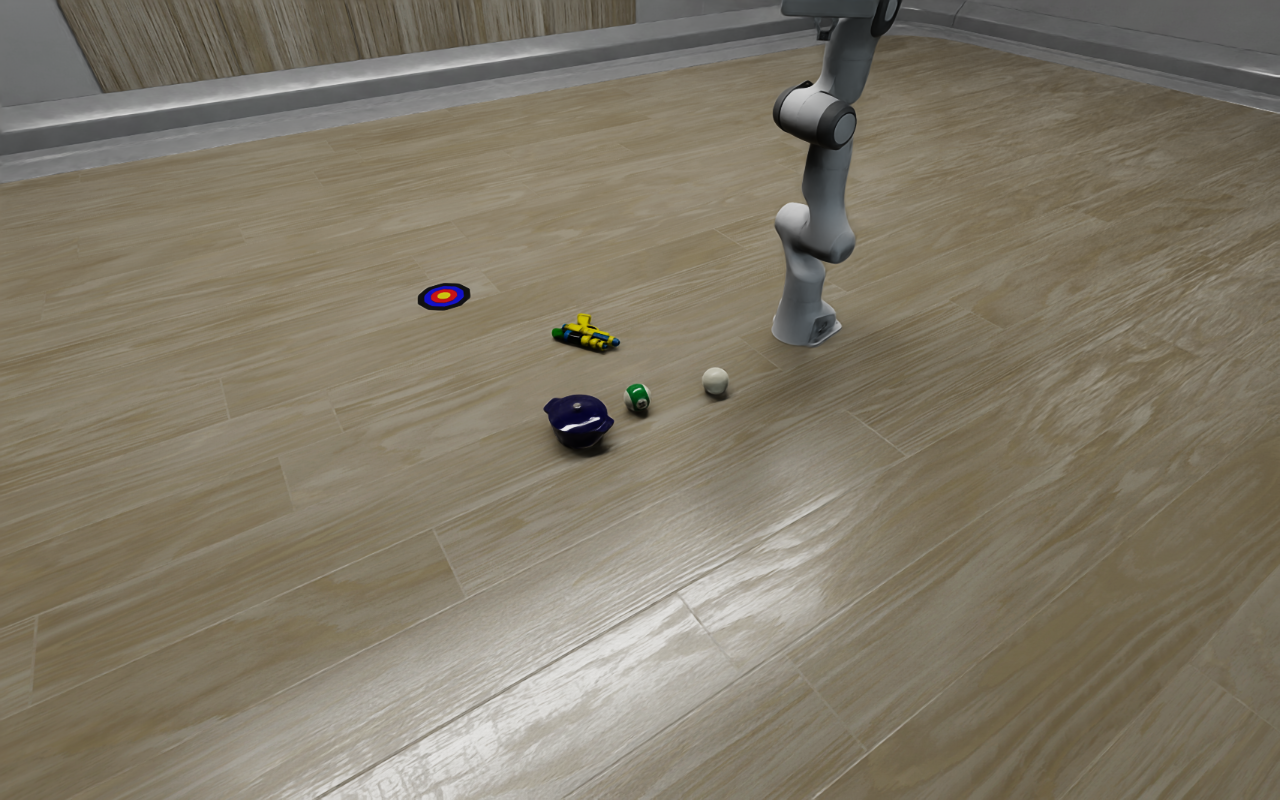} \\
            \noalign{\vskip 1pt}
            \includegraphics[width=\imgwidth]{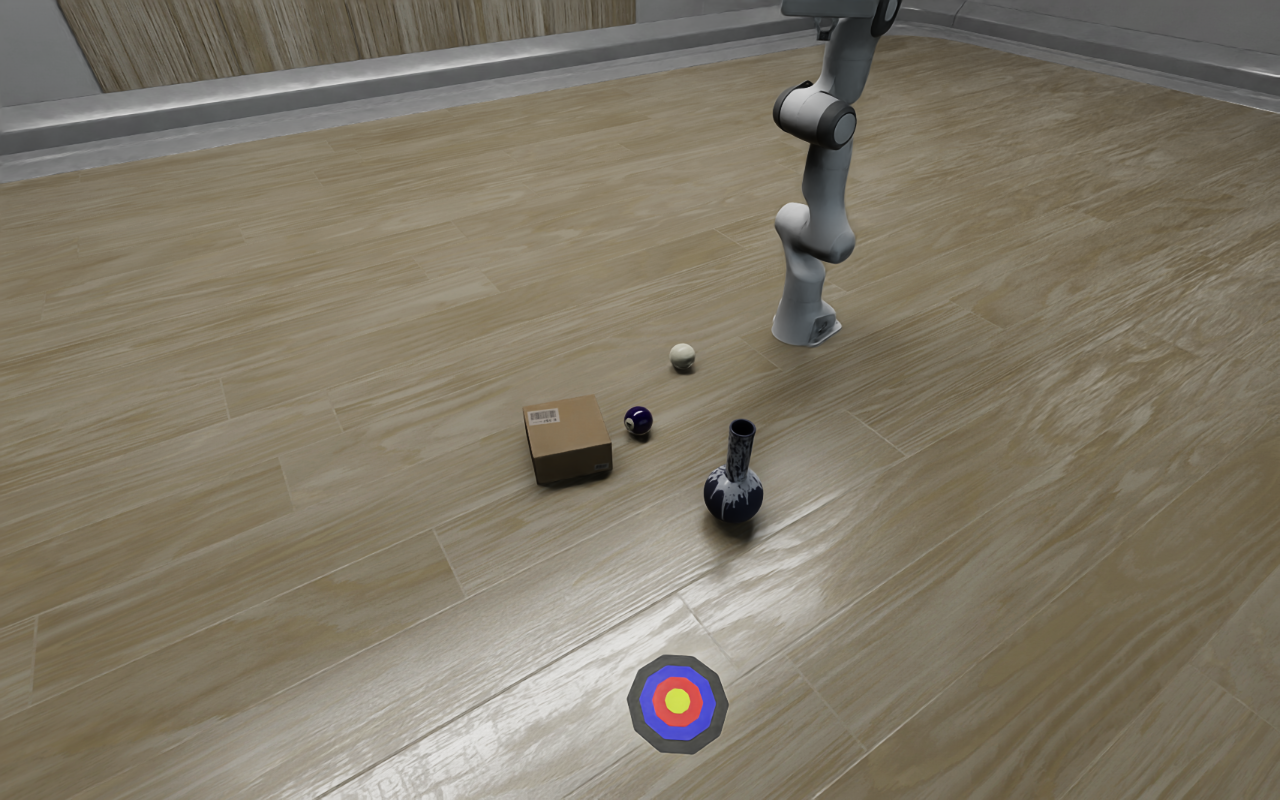} &
            \includegraphics[width=\imgwidth]{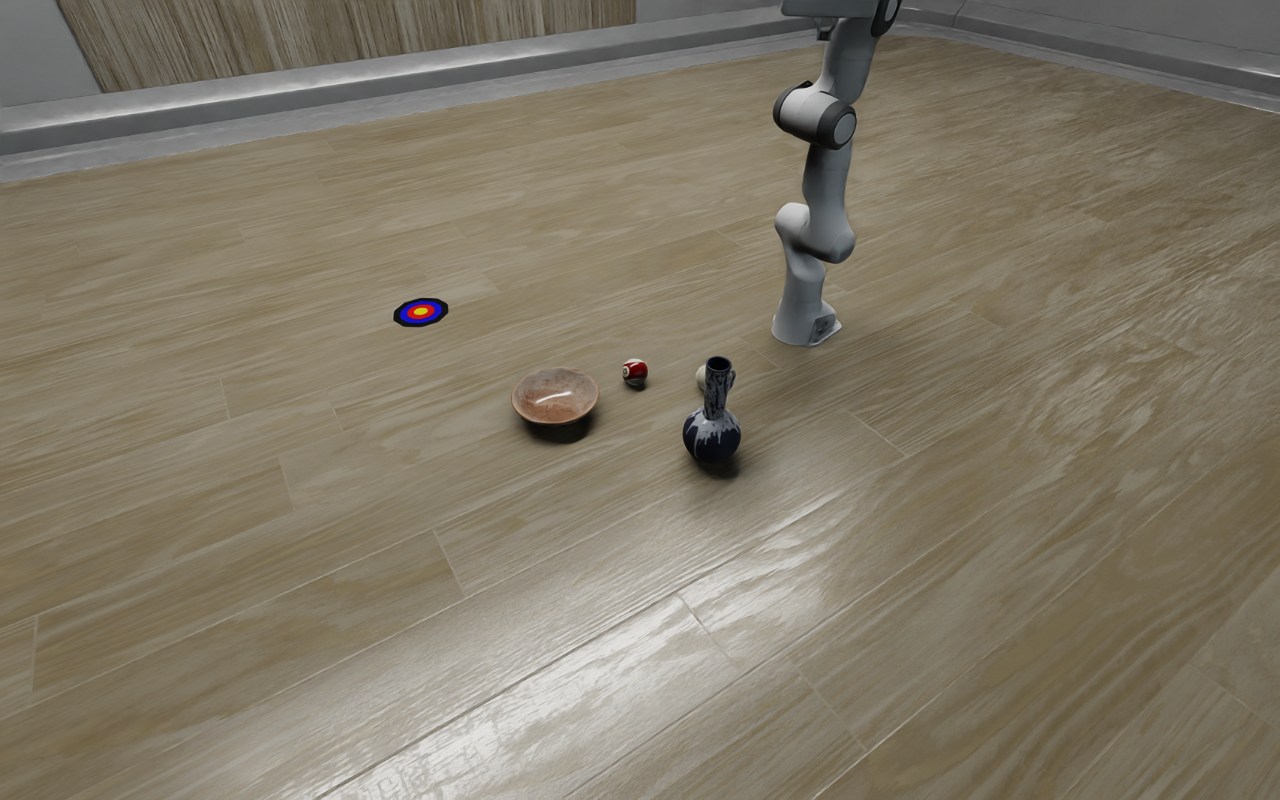} &
            \includegraphics[width=\imgwidth]{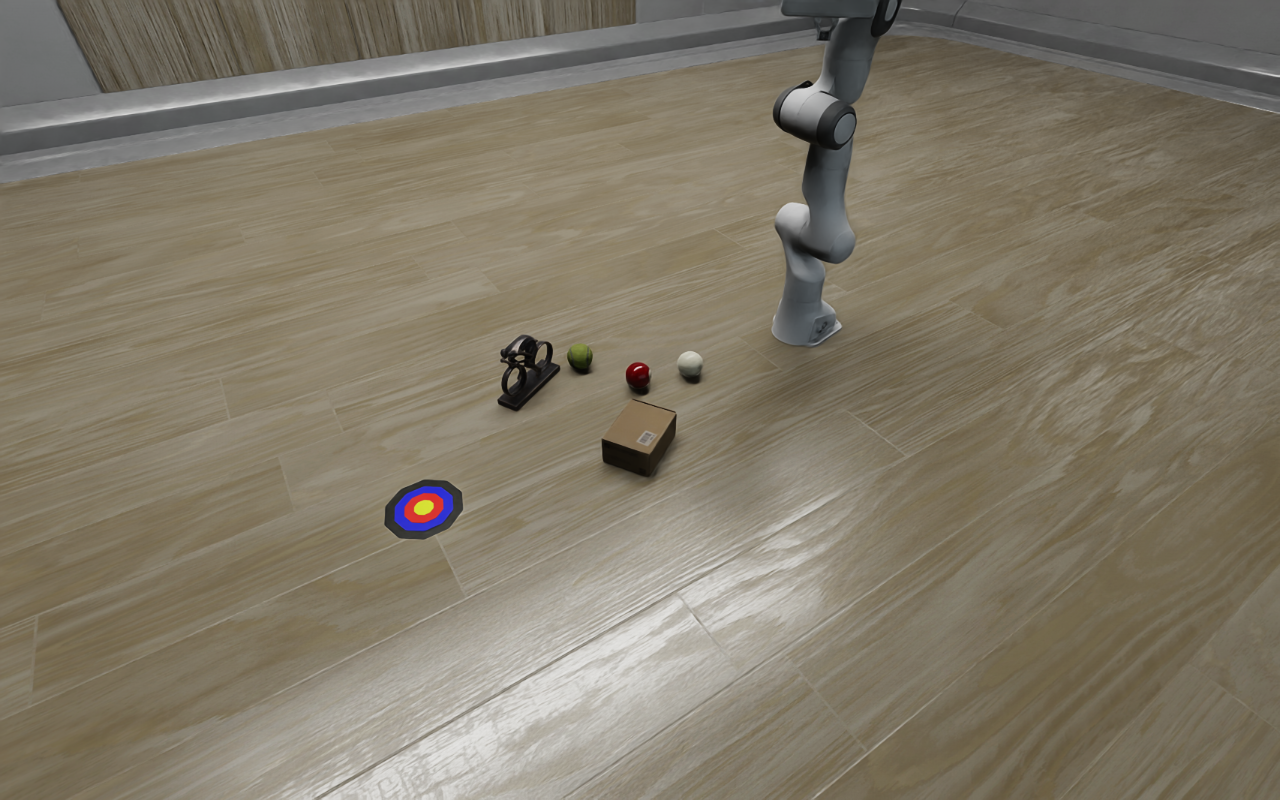} \\
            \noalign{\vskip 1pt}
            \includegraphics[width=\imgwidth]{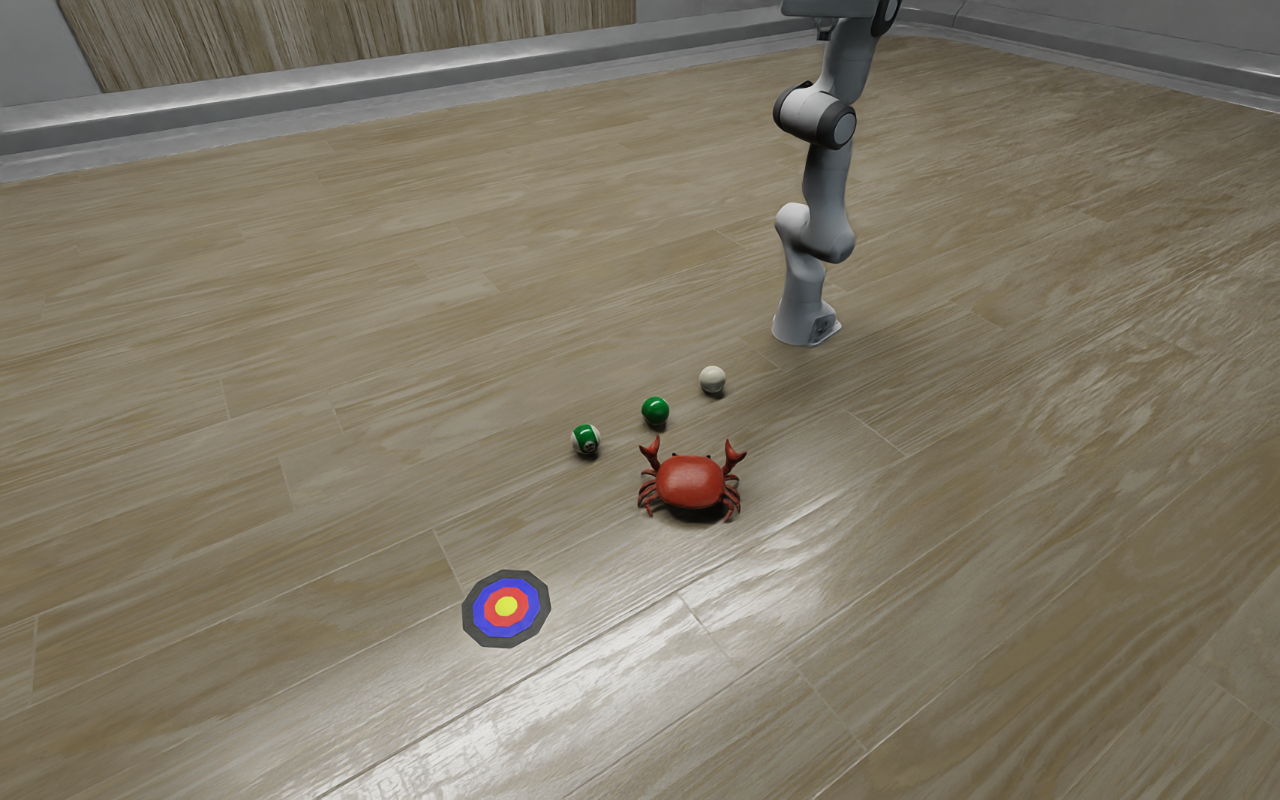} &
            \includegraphics[width=\imgwidth]{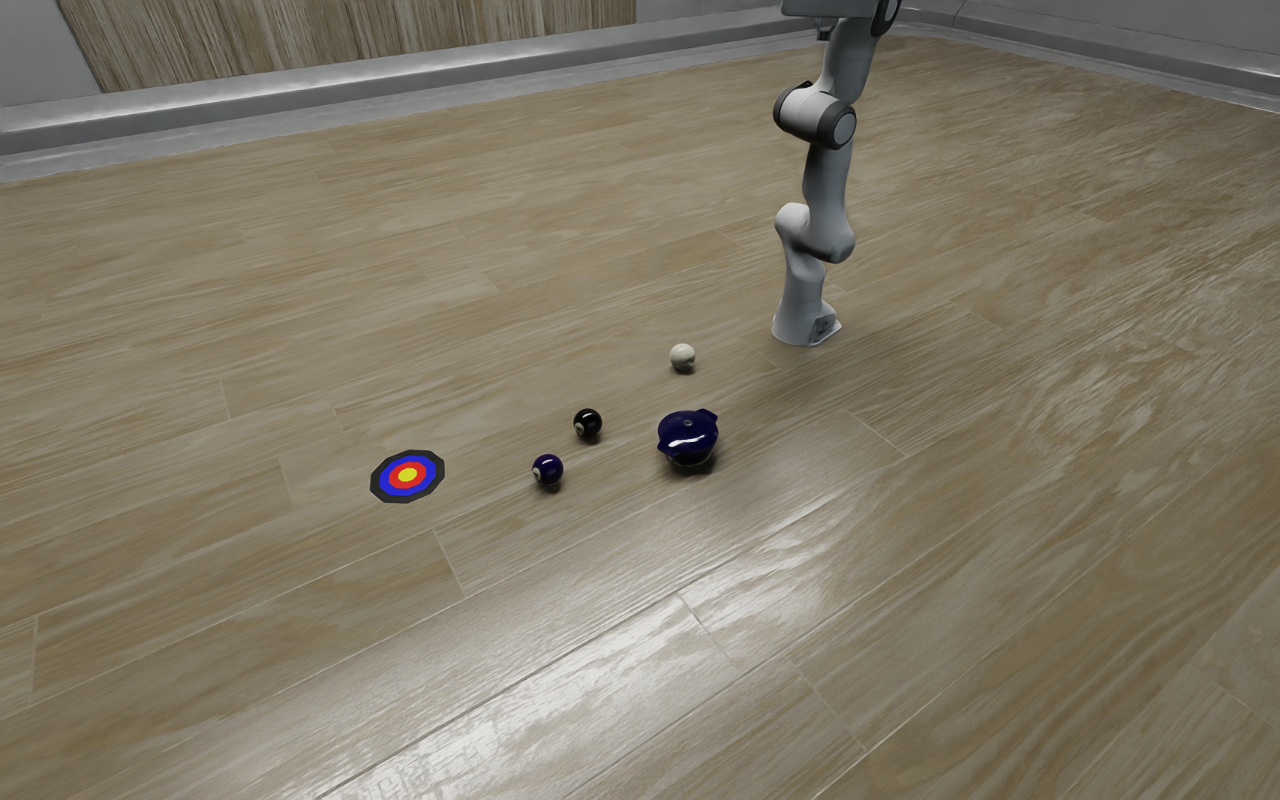} &
            \includegraphics[width=\imgwidth]{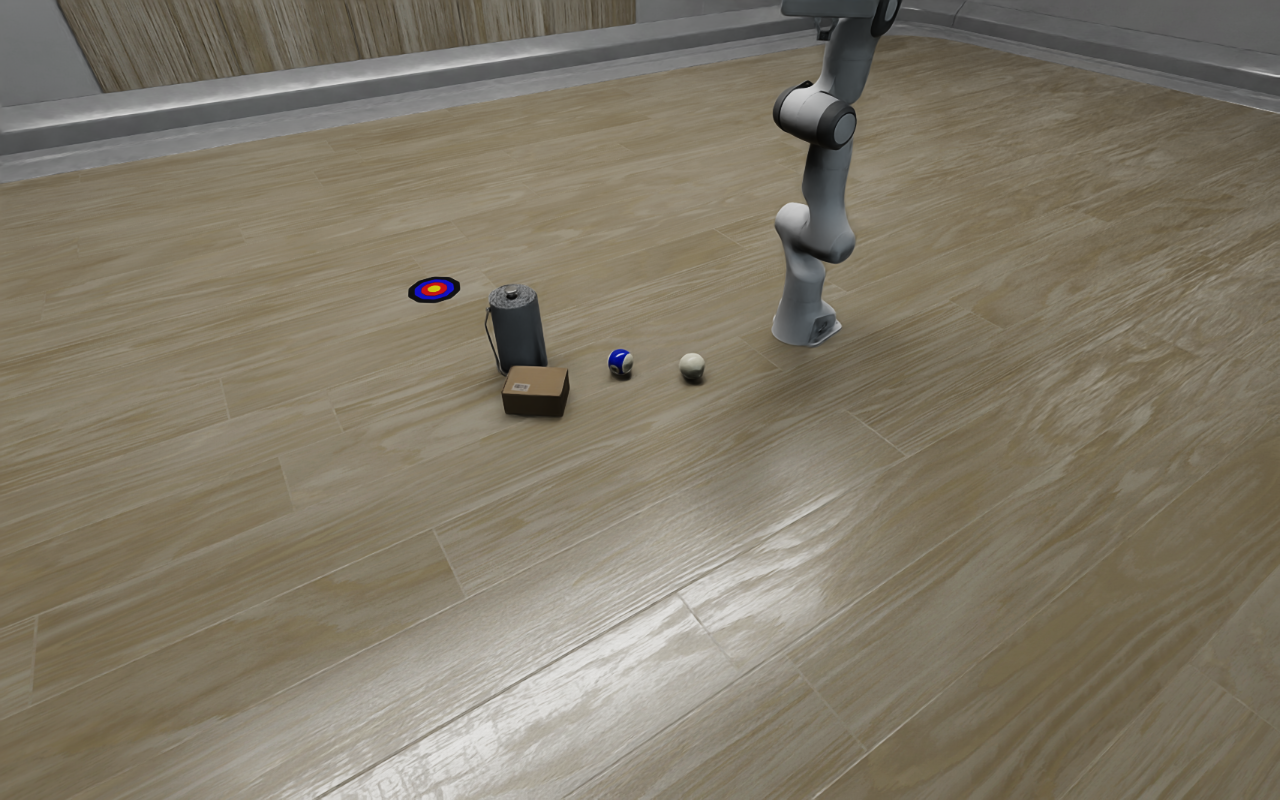} \\
            \noalign{\vskip 1pt}
            \includegraphics[width=\imgwidth]{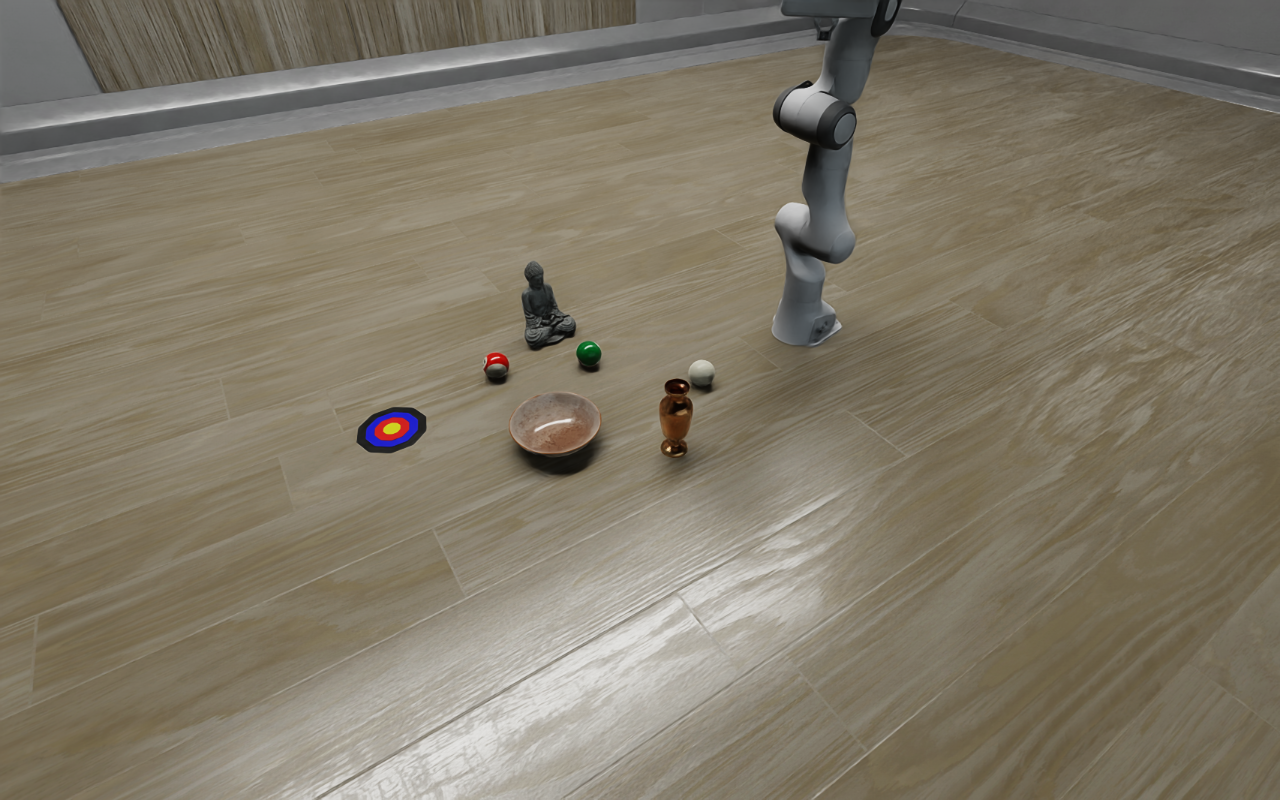} &
            \includegraphics[width=\imgwidth]{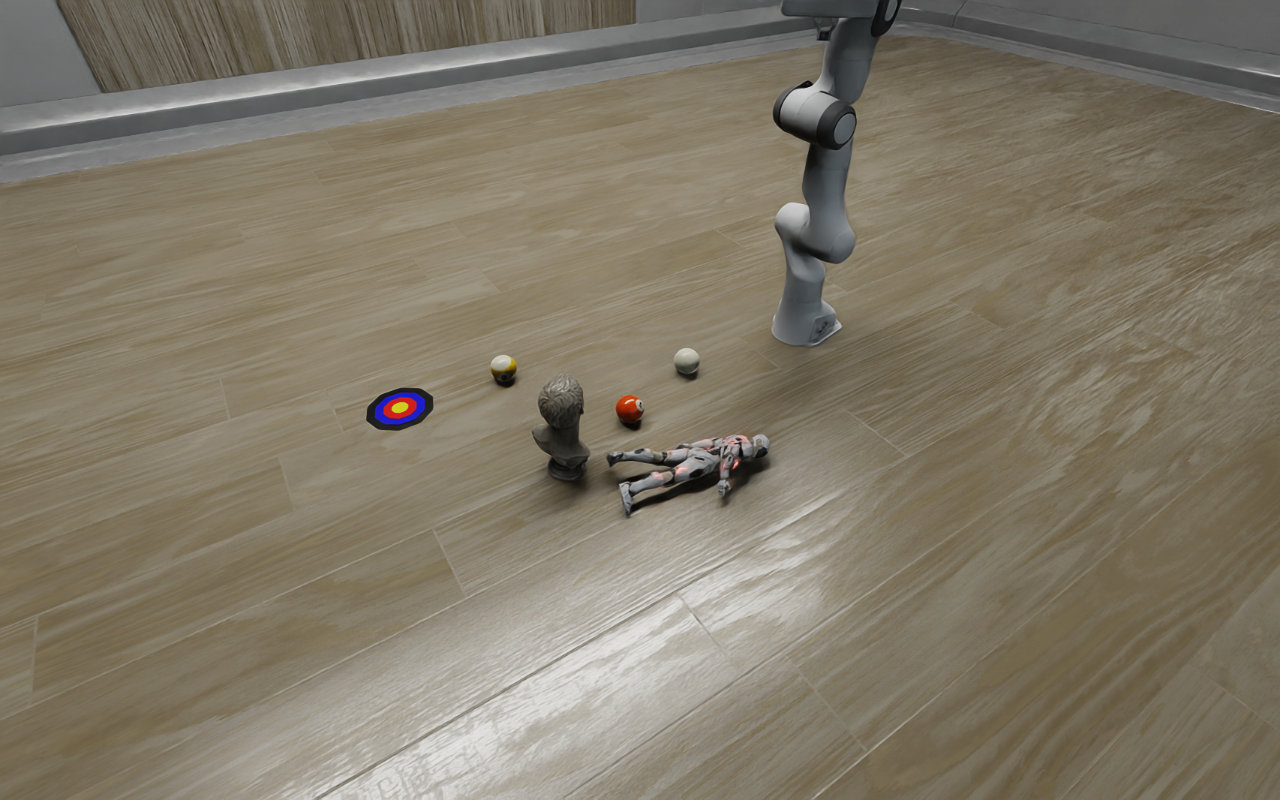} &
            \includegraphics[width=\imgwidth]{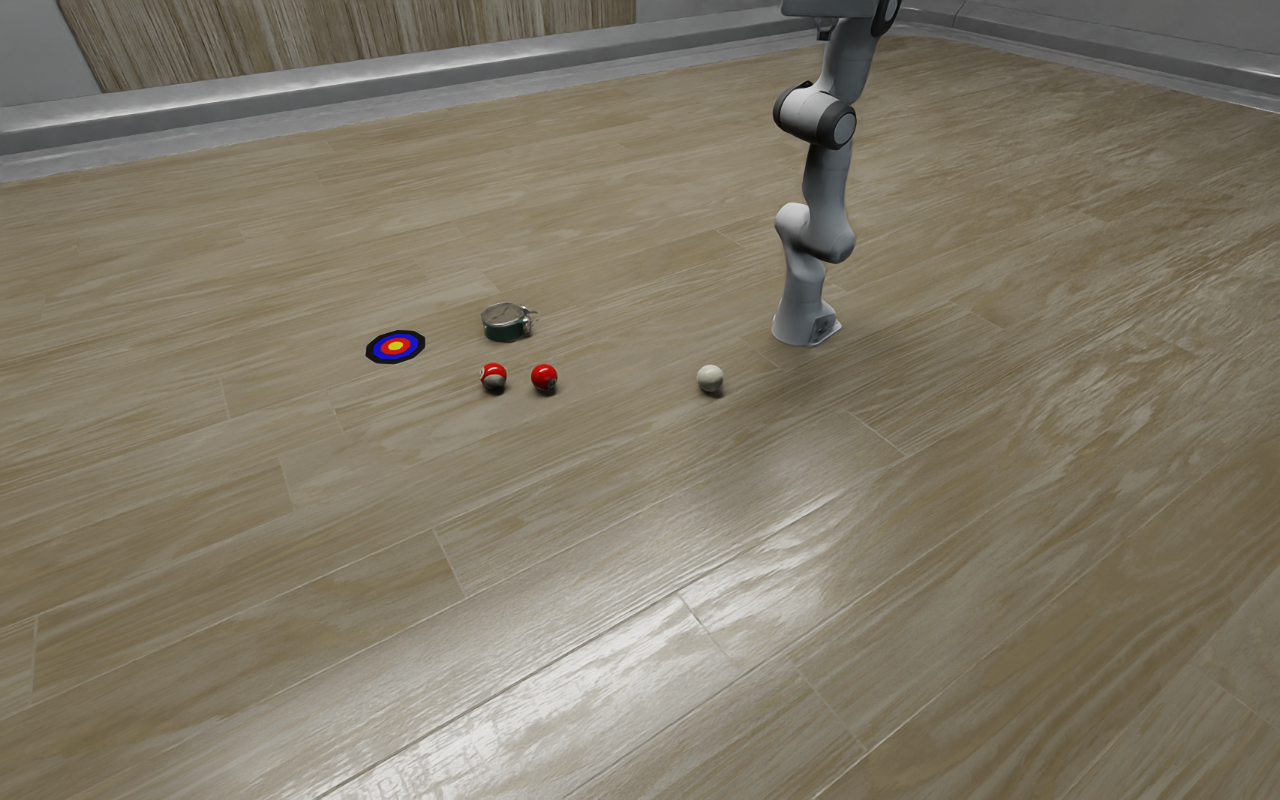}
        \end{tabular}
        \caption{Visualizations of the 18 billiards scenes evaluated in our experiments.}
        \label{fig:appendix_manipulator_scenes}
    \end{figure}

\vspace{-0.1in}
    For our billiards experiment, we generated 18 scenes with the assistance of a procedural generation tool. This tool was written to randomly place a cue ball, a target ball, and a goal position. In all scenes but the first, random objects are then sampled and scattered throughout the scene. After this step, elements of the scene were manually adjusted to ensure that the target ball could reasonably achieve the goal position. We show our full set of environments in \Cref{fig:appendix_manipulator_scenes}.

\vspace{-0.1in}
\label{sec:appendix_manipulator_imaging}
\subsection{Scene Imaging and Reconstruction}
\vspace{-0.1in}
    For each billiards scene, we acquire 60 RGBD images at a resolution of $1280 \times 800$ pixels using a simulated Orbbec Gemini 2 camera. The camera begins at a frontal viewpoint capturing the full workspace, then moves outward along a vertical plane that faces the workspace while sampling viewpoints. Upon reaching the surface of a sphere of 1-meter radius, centered at the manipulator base, the camera continues its trajectory along the hemisphere. The camera continuously repeats this path three times, with its height gradually increasing from 0.1 to 0.25 meters as it moves to ensure complete coverage of the scene objects. The camera is always pointed at the workspace center.
    
    Object detection is performed on the first frame at native resolution to maximize detection accuracy with OWLv2~\citep{minderer2023scaling}. Detected bounding boxes are used to prompt the SAM 2~\citep{ravi2024sam2} segmentation model. Occasionally, predicted masks slightly over- or under-segment object boundaries by a few pixels. To address this, we identify the edges of each segmentation mask and consider all pixels within a 5-pixel margin. For these pixels, we use a distance-based majority vote with k-nearest neighbors in the backprojected point cloud to assign corrected class labels.
    
    For computational efficiency, Gaussian splatting is applied to all RGBD and segmentation images after downsampling to half-resolution ($640 \times 400$): color images are downsampled via cubic interpolation, and depth and segmentation images via nearest-neighbor interpolation.
    
    To obtain material properties, we iteratively query OpenAI's GPT-4o VLM with an image of the workspace with the entity of interest annotated with a bounding box and a text prompt. We provide the prompt template on the following page.

    \newpage
    \begin{tcolorbox}[colback=gray!5, colback=gray!5, title=Material Query Prompt Template, boxrule=0.5pt, enhanced,]
        \begin{tcolorbox}[colback=yellow!6, colframe=yellow!40!black, title=System Message, boxrule=0.5pt, enhanced, before skip=0pt, after skip=2pt]
            \ttfamily
            You are an assistant to an autonomous robot. Your job is to interpret the robot's visual observations and answer its questions. The robot will provide its input query and relevant context under "Input." It will provide more specific requirements pertaining to the query under "Task." Please provide your response in the specified format.
        \end{tcolorbox}
        \begin{tcolorbox}[colback=blue!7, colframe=blue!40!black, title=User Message,  boxrule=0.5pt, enhanced, before skip=0pt, after skip=0pt]
            \ttfamily
            <Image> \\
            
            Input: \\
            The robot observes a scene and detects an object of interest. The image is provided with a bounding box indicating the object of interest. The robot has a preliminary annotation of the object's identity, and indicates it to be <annotation>, which may or may not be correct. The robot seeks verification of the object's identity and physical property estimation. \\
            
            Task: \\
            Verify the identity of the object inside the bounding box. If the annotation is accurate, confirm it. If it is inaccurate, provide the most appropriate object label. Furthermore, provide a general description of the object's identity, physical appearance, and purpose. \\
            
            The robot requires an estimate of the object's physical properties to calculate and plan for environment interactions. First, determine the most appropriate material for the object. If the object appears to be composed of multiple materials or the material is indiscernible, please provide the most prevalent or representative material. Then, estimate the following physical properties of the material: \\
            
            Density (kg/m$^3$) \\
            Friction Coefficient \\
            Coefficient of Restitution \\
            Young’s Modulus (Pa) \\
            Poisson’s Ratio \\
            
            Please provide a single best numerical estimate for each physical property. The output should be structured as a JSON file, with the following fields: \\
            
            -Annotation Accuracy \\
            -Object Label \\
            -General Description \\
            -Material \\
            -Density \\
            -Friction Coefficient \\
            -Coefficient of Restitution \\
            -Young's Modulus \\
            -Poisson's Ratio
        \end{tcolorbox}
        \label{box:material_prompt}
    \end{tcolorbox}

\newpage
\vspace{-0.1in}
\subsection{Manipulator Control and Action Modeling}
\vspace{-0.1in}
    The FR3 manipulator is controlled with an operational space torque controller~\citep{MortonPavone2025}. Each strike action is parameterized by the cue contact position, contact speed, and strike angle. The action is executed as a linear sweep comprising two 0.1-meter segments: an acceleration phase, where the end effector accelerates to the target speed at the contact point, and a deceleration phase, wherein it slows to rest after contact. This controller was used to compute torque commands at a rate of 400 Hz.

\vspace{-0.1in}
\subsection{Virtual Environment}
\vspace{-0.1in}
    We use PyBullet~\citep{coumans2020} as our virtual physics environment and simulate dynamics with a time step of 0.0025 seconds. PyBullet only supports collision detection on convex meshes, therefore we perform a convex decomposition on our meshes using~\citep{wei2022coacd} before loading our reconstructed objects into the simulator. PyBullet simulations were run exclusively on the CPU.

\vspace{-0.1in}
\subsection{Baseline Implementation Details}
\vspace{-0.1in}
    The baseline planner conducts an exhaustive grid search over candidate strike parameters at the cue ball contact point. Strike speed is discretized into 20 evenly spaced values ranging from 0.2 to 0.85~m/s, while strike angle is sampled at 60 increments from $-10^\circ$ to $+10^\circ$ relative to the vector connecting the centers of the cue and target balls. The planner assumes that, immediately after contact, the cue ball acquires the specified strike velocity and angle. For each candidate action, we analytically predict the immediate target ball post-collision state, modeling the cue–target ball collision as perfectly elastic. Specifically, we assume the normal component of the cue ball transfers to the target ball immediately after the collision.

    To further account for secondary interactions with obstacles, we generate a top-down RGBD map of the workspace from the initial reconstruction by threshold-segmenting obstacles observed with a $z$ coordinate greater than 0.005 meters to create a static occupancy mask. The simulated path of the target ball is traced through the workspace on this map. If a collision with an obstacle or workspace boundary is detected, the velocity of the target ball is reflected about the obstacle’s normal direction, updating only its normal velocity component.
    
    The grid search outputs the action predicted to bring the target ball closest to the goal according to this simplified physical model. This method does not update object states or account for subsequent changes in the environment during execution.

\vspace{-0.1in}
\subsection{Supplementary Results}
\begin{figure}[h!]
        \centering
        \includegraphics[width=0.8725\linewidth]{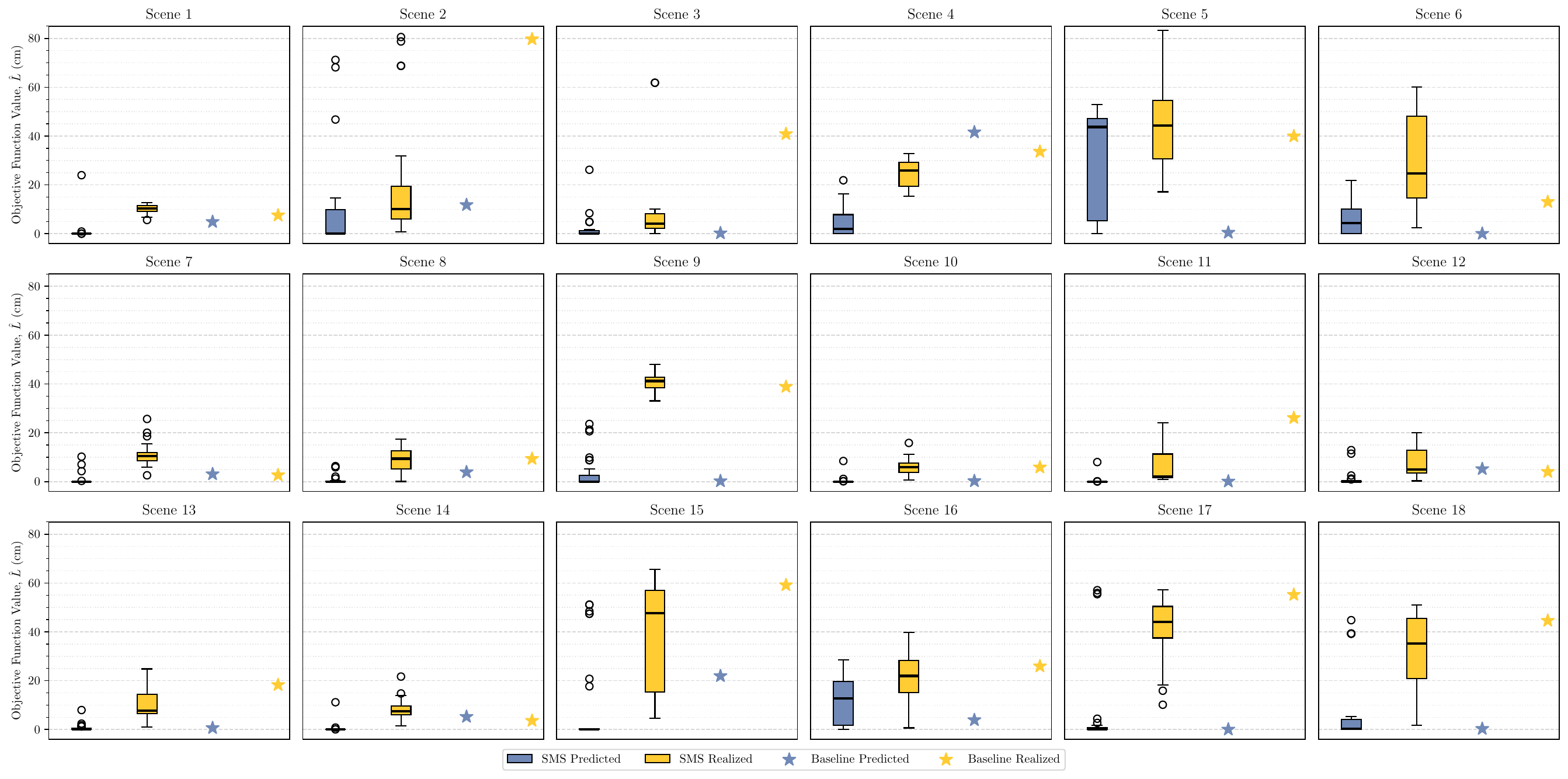}
        \caption{Distributions of SMS performance over 30 repeated action optimizations. Baseline results are shown for comparison. Lower is better. For correspondence with \Cref{fig:experiment_manipulator_dists}, Scenes A, B, C, and D respectively correspond to Scenes 10, 12, 16, and 5 here.}
        \label{fig:appendix_manipulator_dists}
    \end{figure}

\vspace{-0.1in}
    For completeness, an extended version of \Cref{fig:experiment_manipulator_dists} reporting predicted and realized performance results for each individual scene is also included as \Cref{fig:appendix_manipulator_dists}.

\newpage
\section{Additional Details for Quadrotor Landing Scenario}
\renewcommand{\thefigure}{B.\arabic{figure}}
\setcounter{figure}{0}

\vspace{-0.1in}
\subsection{Scene Generation}
\begin{figure}[htbp]
        \newcommand{\imgwidth}{0.5\linewidth}
        \centering
        \setlength{\tabcolsep}{0.5pt}
        \renewcommand{\arraystretch}{0.0}
        \begin{tabular}{cc}
            \includegraphics[width=\imgwidth]{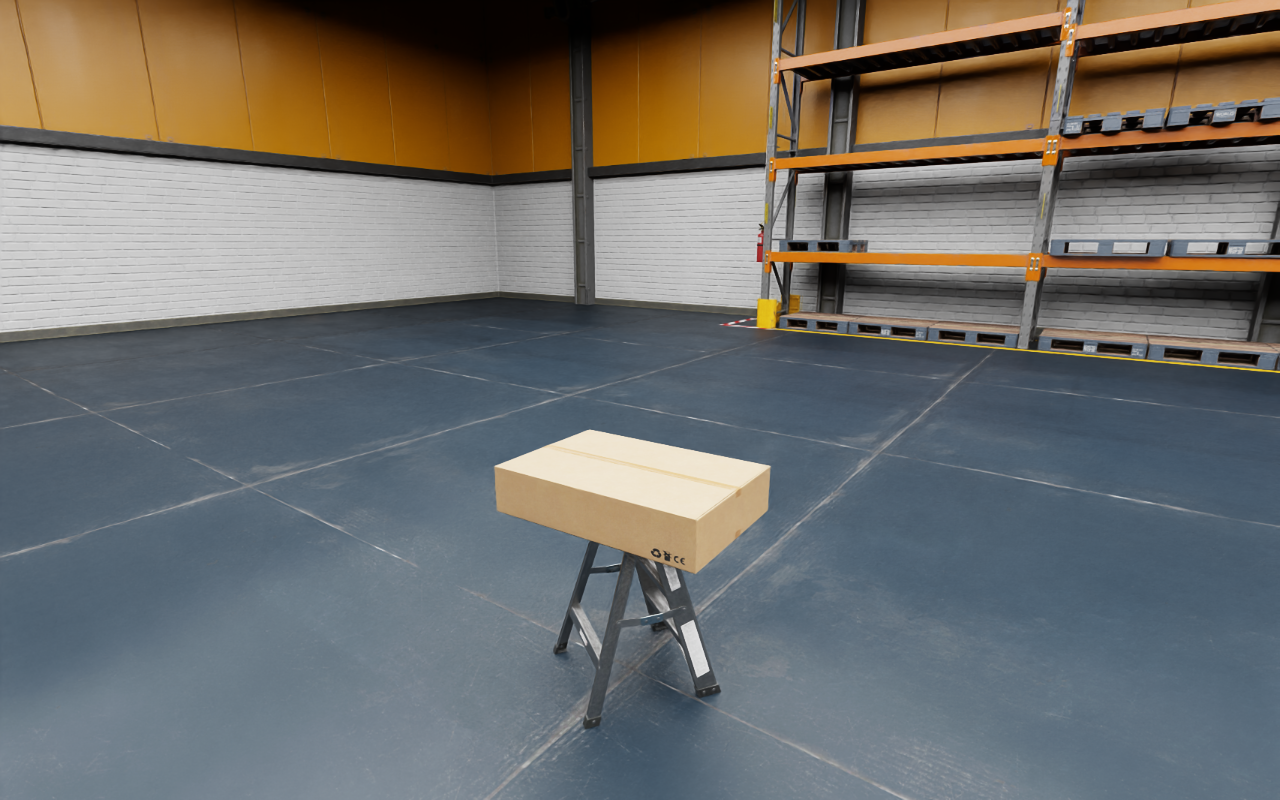} &
            \includegraphics[width=\imgwidth]{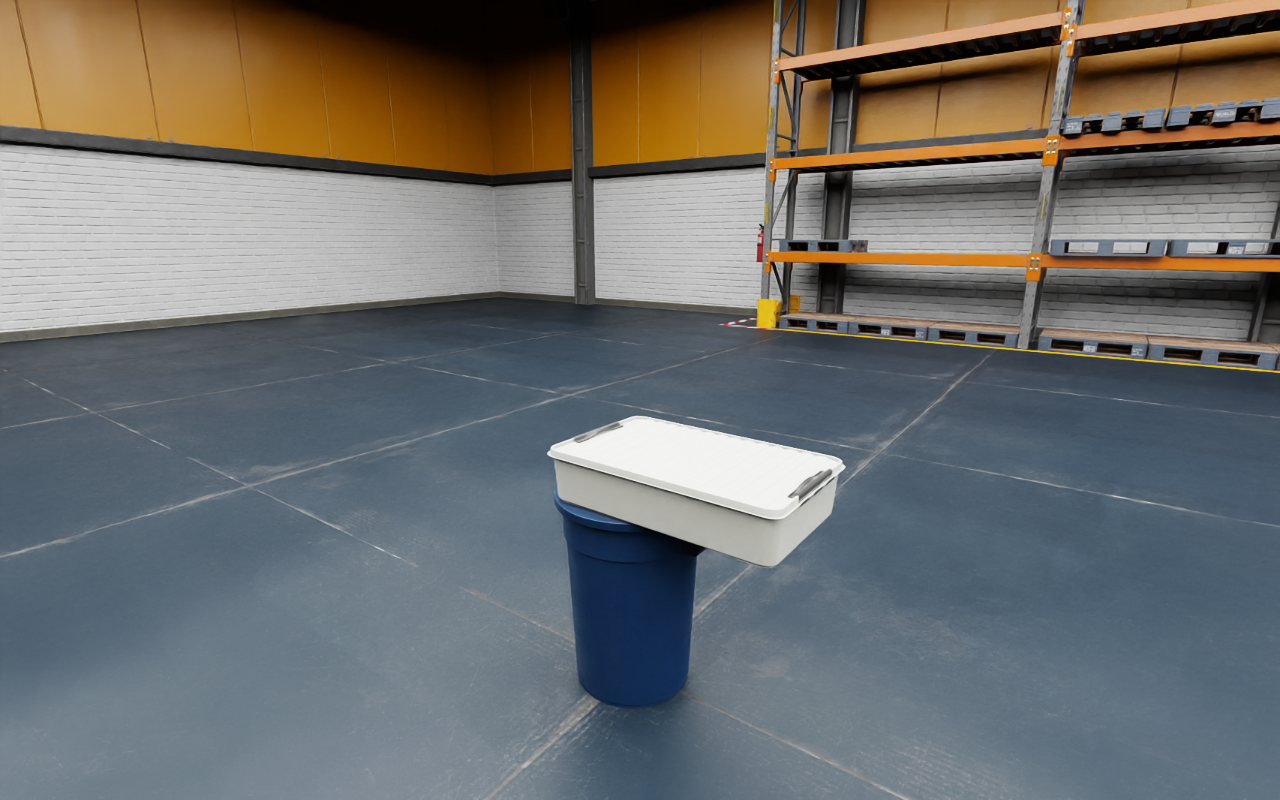} \\
            \noalign{\vskip 1pt}
            \includegraphics[width=\imgwidth]{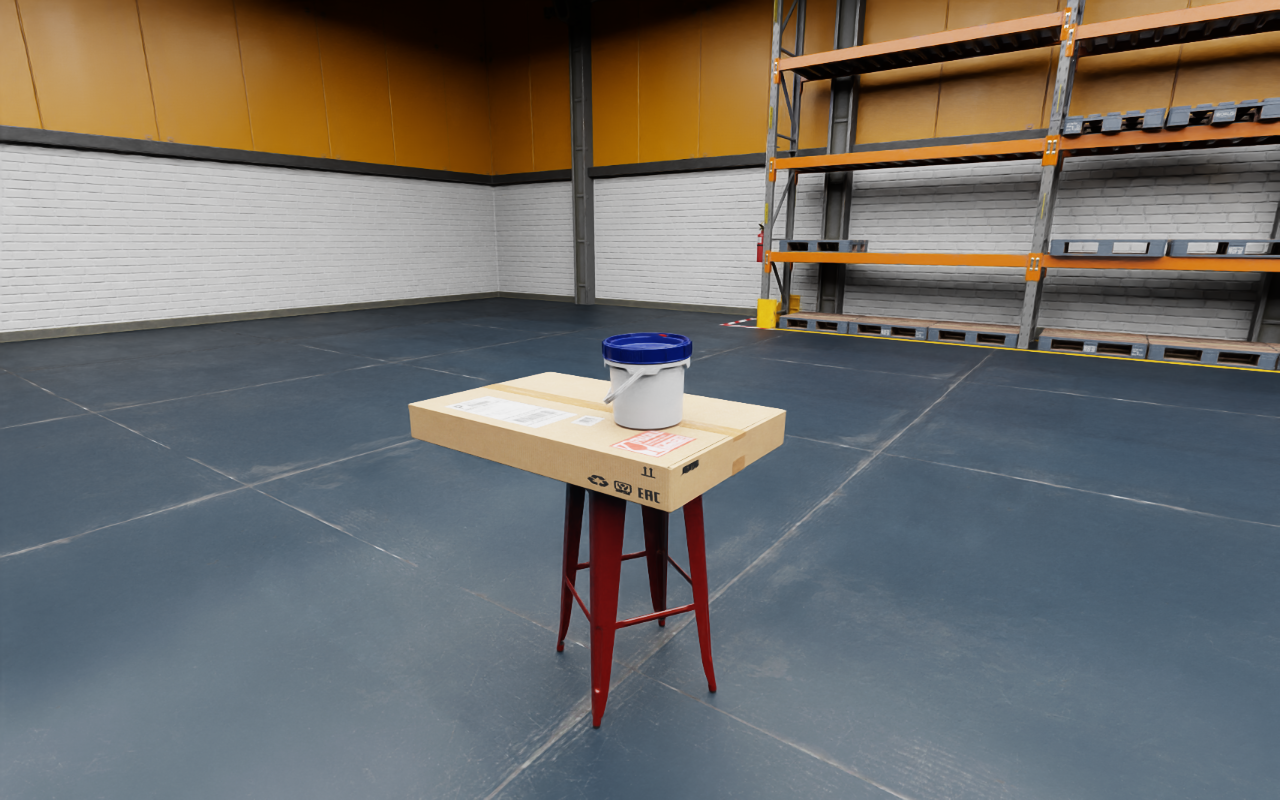} &
            \includegraphics[width=\imgwidth]{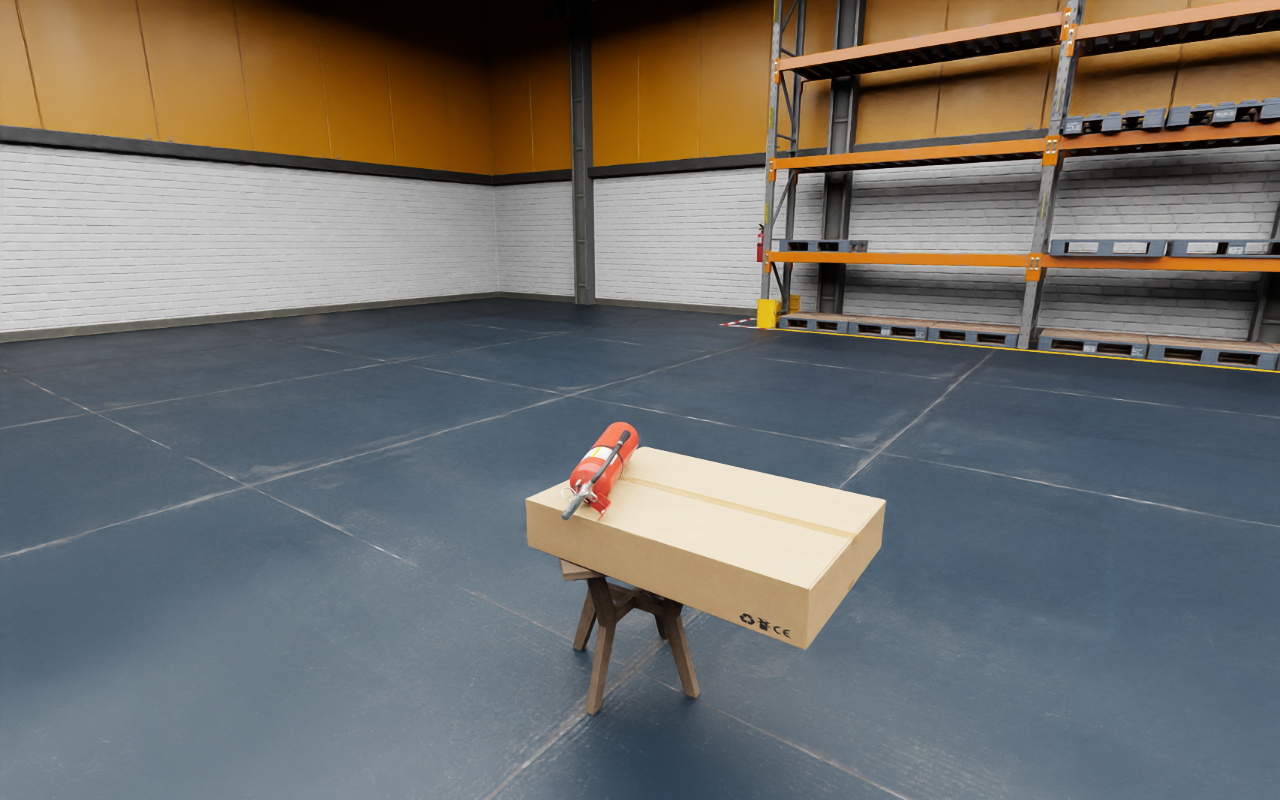} \\
        \end{tabular}
        \caption{Visualizations of the four quadrotor scenes evaluated in our experiments.}
        \label{fig:appendix_quadrotor_scenes}
    \end{figure}

\vspace{-0.1in}
    For the quadrotor landing scenario, we manually constructed four landing structures, each comprising a base and an overhanging landing platform. The base objects were chosen for their varied materials and geometries, while the platforms were selected as flat objects sufficiently large to accommodate the quadrotor. In each scene, the platform was positioned precariously with significant overhang. In two of the scenarios, we further cantilevered the platform by placing a heavy object as ballast, allowing us to evaluate landing behavior under different stability and load conditions. We show our full set of environments in \Cref{fig:appendix_quadrotor_scenes}.

\vspace{-0.1in}
\subsection{Scene Imaging and Reconstruction}
\vspace{-0.1in}
    Imaging and scene reconstruction were performed exactly as in \Cref{sec:appendix_manipulator_imaging}, except that in this scenario, the camera follows a helical path around the landing structure. The camera begins at a height of 0.35 meters and ascends to 1.35 meters, capturing 60 RGBD images over three orbits. Throughout, the camera remains focused on the center of the structure.

\vspace{-0.1in}
\subsection{Quadrotor Propeller Model}
\vspace{-0.1in}
    To simulate realistic propeller downwash effects in the quadrotor landing experiments, we model each propeller as an ideal actuator disk using principles from classical momentum disk theory. This approach allows us to analytically estimate the induced airflow velocity and mass flow rate generated by each propeller, and to calibrate the smoothed-particle hydrodynamics (SPH) emitter parameters accordingly.

    \paragraph{Momentum Disk Theory Overview:}
    For a hovering quadrotor, each propeller must generate a thrust
    \begin{equation*}
        F_\text{prop} = \frac{m_\text{quadrotor} \cdot g}{n_\text{propellers}},
    \end{equation*}
    where $m_\text{quadrotor}$ is the vehicle mass, $g$ is gravitational acceleration, and $n_\text{propellers}$ is the number of propellers.
    
    The propeller is modeled as an ideal disk of area $A_\text{prop} = \pi r_\text{prop}^2$. The actuator disk accelerates the air from a velocity $v_0$, typically approximated as 0 m/s for hovering or low-speed flight, up to $v_\text{exit}$ far after the propeller. According to momentum theory, the thrust is related to the disk-induced velocity by:
    \begin{equation*}
        F_\text{prop} = \dot{m} \cdot (v_\text{exit} - v_0) = \rho_\text{air} A_\text{prop} v_\text{prop} (v_\text{exit} - v_0),
    \end{equation*}
    where $\dot{m}$ is the mass flow rate, $\rho_\text{air}$ is the density of air, and $v_\text{prop}$ is the average velocity through the disk.

    We use Bernoulli's equation to relate the pressure and velocity before and after the propeller disk. The total pressure ahead of the disk is the sum of the static pressure, $P_0$, and dynamic pressure term, $0.5 \rho_\text{air} v_0^2$, as
    \begin{equation*}
        P_{0, \text{total}} = P_0 + 0.5 \rho_\text{air} v_0^2,
    \end{equation*}
    while downstream of the disk the static pressure is
    \begin{equation*}
        P_{\text{exit}, \text{total}} = P_0 + 0.5 \rho_\text{air} v_\text{exit}^2.
    \end{equation*}
    This model predicts a pressure difference at the disk of
    \begin{equation*}
        \Delta P = P_{\text{exit}, \text{total}} - P_{0, \text{total}}.
    \end{equation*}
    As such,
    \begin{equation*}
        F_\text{prop} = \Delta P A_\text{prop} = 0.5 \rho_\text{air} A_\text{prop} (v_\text{exit}^2 - v_0^2).
    \end{equation*}
    Equating the two equations for $F_\text{prop}$ yields $v_\text{prop} = 0.5(v_\text{ext} + v_0)$.
    
    Thus, for hovering ($v_0 = 0$), the induced velocity through the disk is $v_\text{prop} = 0.5 (v_0 + v_\text{exit}) = 0.5 v_\text{exit}$, with $v_0 = 0$, and
    \begin{equation*}
        v_\text{exit} = \sqrt{ \frac{F_\text{prop}}{0.5 \rho_\text{air} A_\text{prop}} }.
    \end{equation*}
    
    \paragraph{Mass Flow Rate and Particle Emitter Calibration:}
    The mass flow rate through the propeller is
    \begin{equation*}
    \dot{m} = \rho_\text{air} A_\text{prop} v_\text{prop}.
    \end{equation*}
    
    To represent this in the SPH simulation, each emitter produces $n_\text{particles}$ cylindrical fluid particles per time step $\Delta t$, each with length $\ell_\text{particle}$ and diameter $d_\text{particle}$. The particle volume is
    \begin{equation*}
    V_\text{particle} = \pi \left( \frac{d_\text{particle}}{2} \right)^2 \ell_\text{particle}.
    \end{equation*}
    The required effective density for each simulated particle to match the physical mass flow is then
    \begin{equation*}
    \rho_\text{particle} = \frac{ \dot{m} \Delta t }{ n_\text{particles} V_\text{particle} }.
    \end{equation*}
    where $\Delta t$ is the simulation time step.
    
    Each emitter in the simulation is thus configured with particles emitted downward ($-z$ direction), using $v_\text{prop}$ as the particle velocity and $\rho_\text{particle}$ as the particle density, so that the total airflow and momentum closely match the theoretical estimate from the real propeller.

\begin{figure}[ht!]
        \centering
        \includegraphics[width=.75\linewidth]{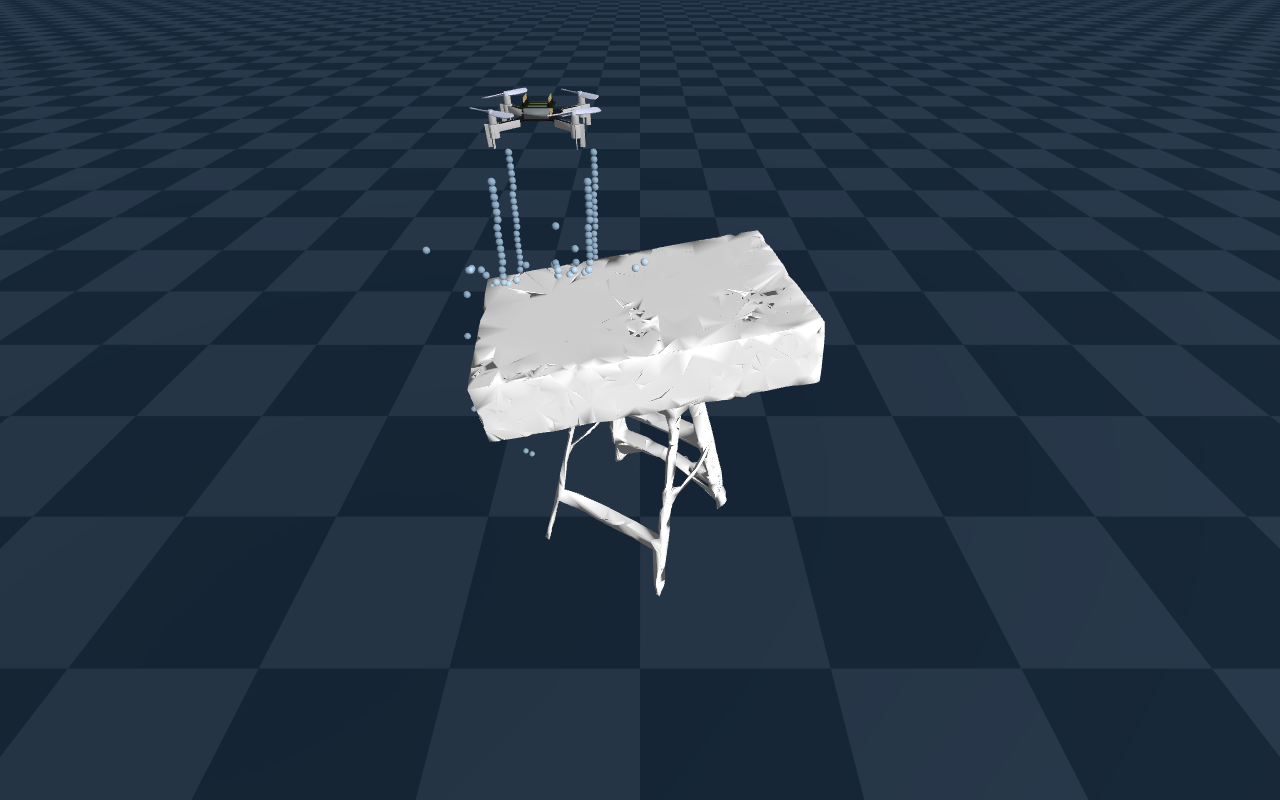}
        \caption{Simulation of propeller downwash in Genesis~\citep{Genesis}, with fluid particles modeling the airflow from each propeller and its impact on the environment, as illustrated by the tilting box.}
        \label{fig:appendix_propeller_sim}
    \end{figure}

    In our simulation, we use $m_\text{quadrotor} =$ 1.182 kg, $r_\text{prop} = 0.0685$ , $\rho_\text{air} =$ 1.225 kg/m$^3$, and, of course, $n_\text{propellers} =$ 4. Other particle parameters such as $n_\text{particles}$, $\ell_\text{particles}$, and $d_\text{particles}$ are adaptively determined by the simulator as a function of time step and emission velocity. We show an image of the quadrotor's virtual environment with simulated propeller fluid particle emissions in \Cref{fig:appendix_propeller_sim}.

\vspace{-0.1in}
\subsection{Virtual Environment}
\vspace{-0.1in}
    Genesis~\citep{Genesis} is used as the virtual physics environment, with dynamics simulated at a 0.01-second time step. To enable efficient fluid particle simulation, all computations are performed on the GPU.

\vspace{-0.1in}
\subsection{Baseline Implementation Details}
\begin{figure}[h!]
        \newcommand{\imgwidth}{0.4\linewidth}
        \centering
        \setlength{\tabcolsep}{0.5pt}
        \renewcommand{\arraystretch}{0.0}
        \begin{tabular}{cc}
            \includegraphics[width=\imgwidth]{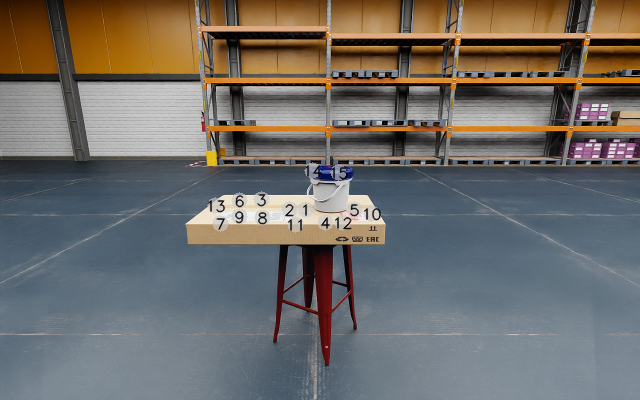} &
            \includegraphics[width=\imgwidth]{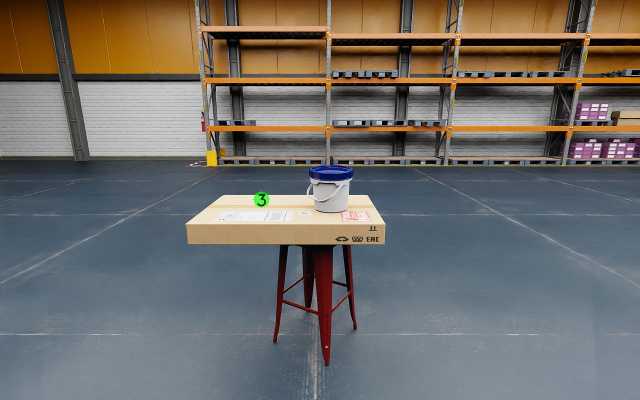} \\
            \noalign{\vskip 1pt}
            \includegraphics[width=\imgwidth]{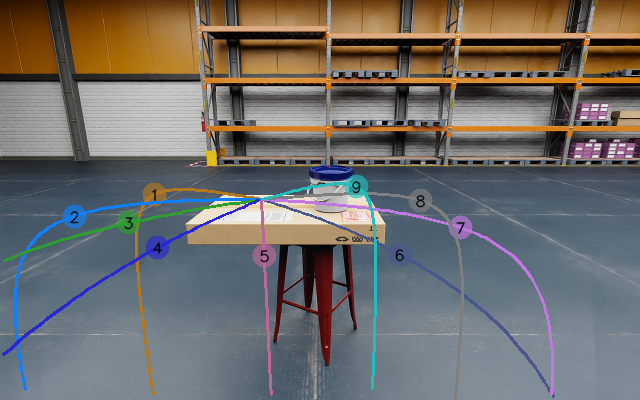} &
            \includegraphics[width=\imgwidth]{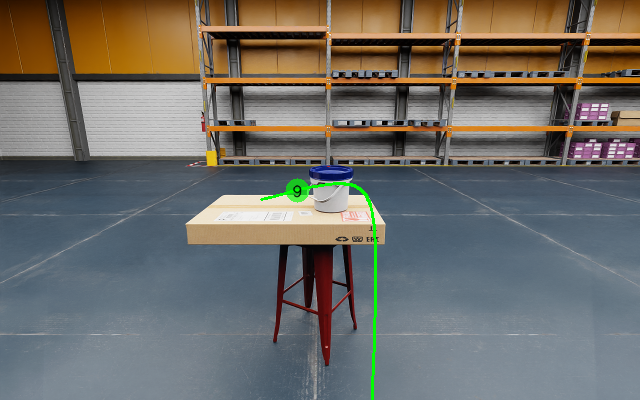} \\
        \end{tabular}
        \caption{Baseline landing site and approach selection. Top left: Candidate landing sites identified by surface geometry and height. Top right: Landing site selected by GPT-4o. Bottom left: Candidate approach paths to the selected site. Bottom right: Final approach chosen by GPT-4o.}
        \label{fig:appendix_quadrotor_baseline}
    \end{figure}

\vspace{-0.1in}
    For the quadrotor baseline, we derive candidate landing sites and approach trajectories using a combination of geometric analysis and visual prompting. Starting with the quadrotor’s RGBD observation, we extract a point cloud and identify surface points with an approximately vertical normal and a height of at least 0.5 meters above the ground, which filters out ground-level points. Up to 15 non-overlapping candidate landing sites are then sampled from these points, ensuring spatial diversity; each site is marked as a circle on the image and assigned a unique identifier. The annotated image is provided as input to GPT-4o, which is prompted to select the most appropriate landing site. Next, we generate 9 candidate approach paths to the selected site, parameterized as Bezier curves and spanning a range of directions. Each path is annotated on the image with its corresponding identifier. GPT-4o is queried again to select the preferred trajectory, yielding the final approach path and landing site for the baseline. An example of these annotated images along with the selected results are shown in \Cref{fig:appendix_quadrotor_baseline}. Prompt templates are also provided in the following pages.

    \begin{tcolorbox}[colback=gray!5, colback=gray!5, title=Landing Position Query Prompt Template, boxrule=0.5pt, enhanced]
        \begin{tcolorbox}[colback=yellow!6, colframe=yellow!40!black, title=System Message, boxrule=0.5pt, enhanced, before skip=0pt, after skip=2pt]
            \ttfamily
            You are an assistant to an autonomous robot. Your job is to interpret the robot's visual observations and answer its questions. The robot will provide its input query and relevant context under "Input." It will provide more specific requirements pertaining to the query under "Task." Please provide your response in the specified format.
        \end{tcolorbox}
        \begin{tcolorbox}[colback=blue!7, colframe=blue!40!black, title=User Message,  boxrule=0.5pt, enhanced, before skip=0pt, after skip=0pt]
            \ttfamily
            <Image> \\
            
            Input: \\

            The robot is a quadrotor looking to land on the <landing\_target>. It needs to land on a stable position that will neither cause it to fall nor topple the landing platform or nearby objects. \\
            
            Task: \\
            
            An image of the quadrotor's current observation is provided. Landing position candidates are provided and are annotated on the image as circles with IDs. Which of these landing sites should the quadrotor choose as its landing position? \\
            
            Please provide the output in the following format: \\
            
            Reasoning: (e.g., What should the quadrotor consider? What are the risks? What are the safe areas?) \\
            Decision: (ID of landing location; Please only specify the ID number)
        \end{tcolorbox}
        \label{box:material_prompt}
    \end{tcolorbox}

    \begin{tcolorbox}[colback=gray!5, colback=gray!5, title=Approach Path Query Prompt Template, boxrule=0.5pt, enhanced]
        \begin{tcolorbox}[colback=yellow!6, colframe=yellow!40!black, title=System Message, boxrule=0.5pt, enhanced, before skip=0pt, after skip=2pt]
            \ttfamily
            You are an assistant to an autonomous robot. Your job is to interpret the robot's visual observations and answer its questions. The robot will provide its input query and relevant context under "Input." It will provide more specific requirements pertaining to the query under "Task." Please provide your response in the specified format.
        \end{tcolorbox}
        \begin{tcolorbox}[colback=blue!7, colframe=blue!40!black, title=User Message,  boxrule=0.5pt, enhanced, before skip=0pt, after skip=0pt]
            \ttfamily
            <Image> \\
            
            Input: \\

            The robot is a quadrotor looking to land on the <landing\_target>. It needs to land on a stable position that will neither cause it to fall nor topple the landing platform or nearby objects. It has identified a landing position and is now attempting to determine the best approach path. The approach path must account for the quadrotor propeller wash, which can impart a force on the objects below it, including the landing platform (i.e., the <landing\_target>). The quadrotor should seek a path that minimally disturbs objects that it will fly over, and should especially try to avoid toppling the landing area. \\
            
            Task: \\
            
            An image of the quadrotor's current observation is provided. Approach path candidates are annotated as curves of different colors, each with a corresponding numerical ID. Note that these paths are projected to the height of the landing platform and the quadrotor would be flying at some small distance above these paths until it arrives at the landing position where it will descend. Which of these approach paths should the quadrotor take? \\
            
            Please provide the output in the following format: \\
            
            Reasoning: (e.g., What should the quadrotor consider? What are the risks?) \\
            Decision: (ID of approach path; Please only specify the ID number)
        \end{tcolorbox}
        \label{box:material_prompt}
    \end{tcolorbox}


\end{document}